\documentclass[letterpaper, 10 pt, conference]{ieeeconf}  

\IEEEoverridecommandlockouts                              

\usepackage[utf8]{inputenc}
\usepackage{mathtools}
\usepackage{latexsym}
\usepackage[hidelinks]{hyperref}
\usepackage{subcaption}
\usepackage{float}
\usepackage{color}
\usepackage[export]{adjustbox}
\usepackage{amsmath}
\usepackage{multicol}
\usepackage{fancyhdr}
\usepackage{graphicx}
\usepackage{wrapfig}
\usepackage{listings}
\usepackage{color}
\usepackage{etoolbox}
\usepackage{float}
\usepackage{caption}
\let\labelindent\relax
\usepackage{enumitem}
\usepackage{amssymb}
\usepackage{algorithm}
\usepackage{algpseudocode}
\usepackage{siunitx}

\usepackage{xcolor} 
\usepackage{hyperref}

\hypersetup{
     colorlinks=true,
     linkcolor=blue,
     filecolor=blue,
     citecolor=green,      
     urlcolor=cyan,
     }

\renewcommand{\v}[1]{{\boldsymbol{\mathbf{#1}}}}
\newcommand\latent{\v{z}}
\newcommand\obs{\v{x}}
\newcommand\rhovec{\v{\rho}}
\newcommand\action{\v{a}}
\newcommand\muvec{\v{\mu}}
\newcommand\g{g}
\newcommand\f{f}
\newcommand\xnoise{\v{r}}
\newcommand\znoise{\v{w}}
\newcommand\joints{\v{q}}

\graphicspath{{Images/}}

\title{\LARGE \bf Multimodal VAE Active Inference Controller}

\author{Cristian Meo$^{1}$ and Pablo Lanillos$^{2}$
\thanks{$^{1}$Cristian Meo is with the Faculty of Mechanical Engineering, Department of Cognitive Robotics, Delft University of Technology, Delft, The Netherlands
        {\tt\small c.meo@student.delft.nl}}%
\thanks{$^{2}$Pablo Lanillos is with the Donders Institute for Brain, Cognition and Behavior, Department of Artificial Intelligence, Radboud University, Nijmegen, The Netherlands.
        {\tt\small p.lanillos@donders.ru.nl}}%
}

\begin{document}

\maketitle

\begin{abstract}
Active inference, a theoretical construct inspired by brain processing, is a promising alternative to control artificial agents. However, current methods do not yet scale to high-dimensional inputs in continuous control. Here we present a novel active inference torque controller for industrial arms that maintains the adaptive characteristics of previous proprioceptive approaches but also enables large-scale multimodal integration (e.g., raw images). We extended our previous mathematical formulation by including multimodal state representation learning using a linearly coupled multimodal variational autoencoder. We evaluated our model on a simulated 7DOF Franka Emika Panda robot arm and compared its behavior with a previous active inference baseline and the Panda built-in optimized controller. Results showed improved tracking and control in goal-directed reaching due to the increased representation power, high robustness to noise and adaptability in changes on the environmental conditions and robot parameters without the need to relearn the generative models nor parameters retuning.
\end{abstract}

\begin{keywords}
Active inference, Bio-inspired perception and action, Learning and adaptive systems, Free energy principle.
\end{keywords}

\section{Introduction}
\label{sec:intro}

Active inference (AIF) is prominent in neuroscientific literature as a general mathematical framework of the brain at the computational level~\cite{friston2010free}. According to this theory, the brain learns by interaction a generative model of the world/body that is used to perform state estimation (perception) as well as to execute actions, optimizing one single objective: Bayesian model evidence. This approach, which grounds on hierarchical variational inference and dynamical systems estimation \cite{friston2008dem}, has strong connections with Bayesian filtering~\cite{sarkka2013bayesian} and control as inference~\cite{millidge2020relationship}, as it both estimates the system state and computes the control commands as a result of the inference process. 

In the last years, some proof-of-concept studies in robotics have shown that AIF may be a powerful framework to address key challenges in robotics, such as adaptation, robustness and abstraction in goal-directed tasks \cite{lanillos2018adaptive, oliver2021empirical, sancaktar2020end, pezzato2020novel, pio2016active, minju2019goal, baioumy2020active, meera2020free, lanillos2020robot}. For instance, a state estimation algorithm and an AIF-based reaching controller for humanoid robots were proposed in \cite{lanillos2018adaptive} and \cite{oliver2021empirical} respectively, showing robust sensory fusion (visual, proprioceptive and tactile) and adaptability to unexpected sensory changes. However, they could only handle low-dimensional inputs. Meanwhile, AIF-based joint torque control was investigated using proprioceptive inputs \cite{pezzato2020novel}, showing better performance than a state-of-the-art model reference adaptive controller. Latterly, we presented a pixel-based deep AIF controller~\cite{sancaktar2020end} that incorporated generative model learning using convolutional neural networks but limited to one sensor modality (visual) and velocity control. Thus, this work investigates how to design a brain-inspired controller, with the adaptability and robustness properties of AIF, that: 1) scales to high-dimensional inputs, 2) allows multisensory integration and 3) commands in torque.

\subsection{Contribution}
\begin{figure}[hbtp!]
	\centering
	\subfloat[Architecture]{
		\centering
		\includegraphics[width=0.8\columnwidth, height=150px]{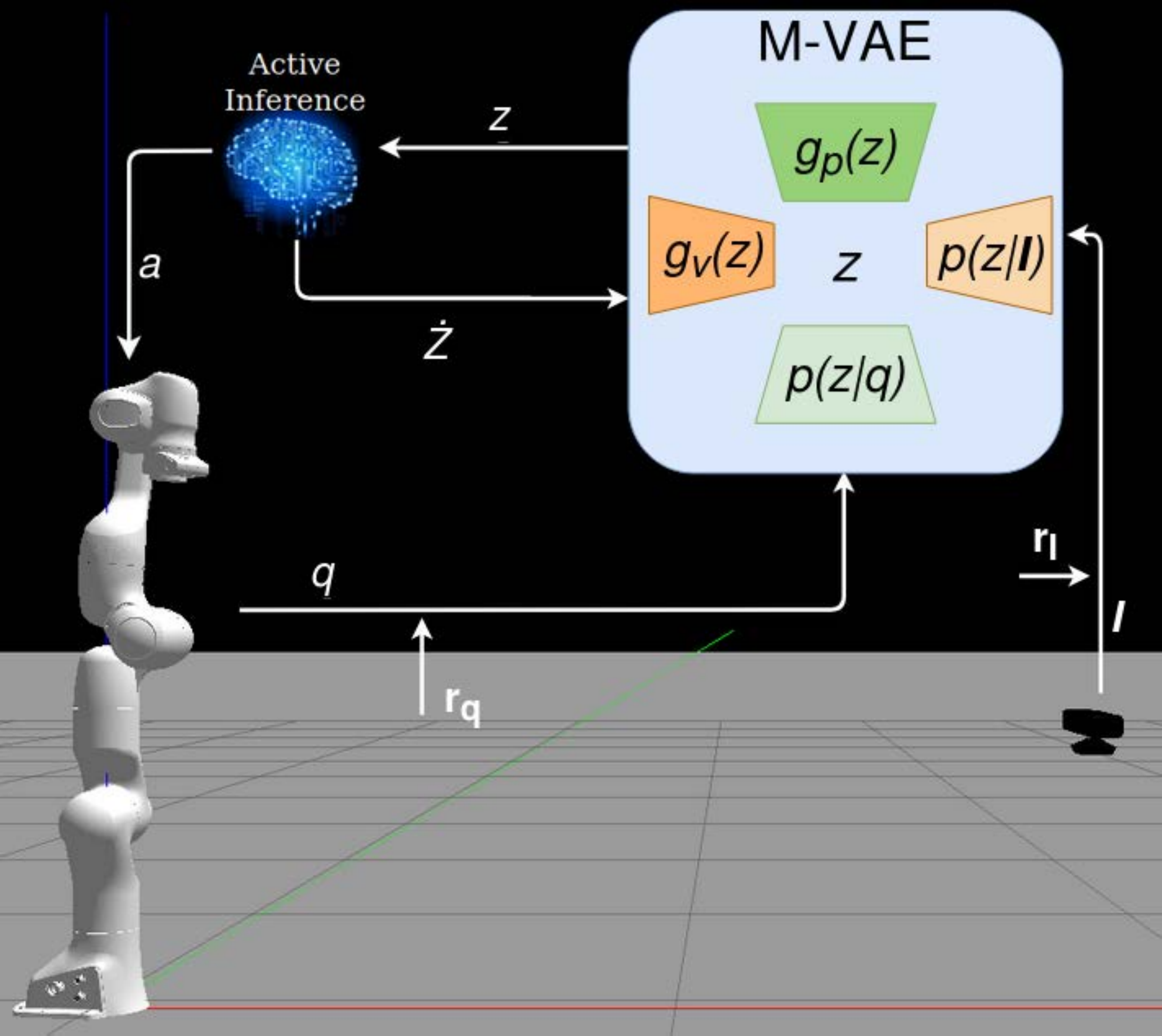}
		\label{fig:architecture:def}
	}\\
	\subfloat[Camera input $\v{I}$]{
	        \includegraphics[width=0.38\columnwidth]{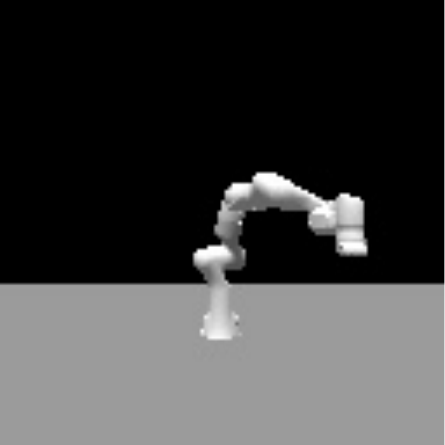}
	  \label{fig:architecture:visual}
    }
    \subfloat[Predicted input $\g_v(\latent)$]{
        \includegraphics[width=0.38\columnwidth]{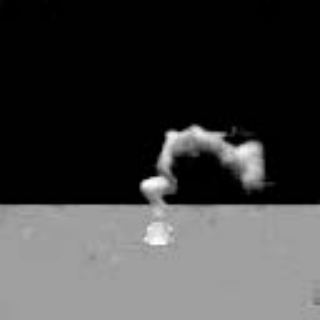}
         \label{fig:architecture:visual_out}
    }
	\caption{MVAE-AIF architecture. (a) Abstract description of the algorithm and environmental setup. State representation learning is given by the multimodal VAE, and estimation and control are provided by the AIF framework. The Panda robot had access to (noisy) proprioceptive input $\joints$ and a camera placed in front $\v{I}$. (b) Camera input. (c) Predicted visual input from the generative model.}
	\label{fig:architecture}
\end{figure}

We propose a Multimodal Vartiational Autoencoder Active Inference (MVAE-AIF) torque controller that combines free energy optimization \cite{friston2010action} with generative model learning (Fig. \ref{fig:architecture}).

We extended our previous AIF formulation \cite{oliver2021empirical, sancaktar2020end} to work with high-dimensional multimodal input at the torque level. Inspired by \cite{Yuge2019Variational}, \cite{sancaktar2020end} and \cite{rood2020deep}, we designed a VAE that provides light coupled multimodal state representation learning \cite{lesort2018state}. One of the main advantages of our approach is that the robot only has to learn the kinematic forward mapping and then is the inference process that produces the right torques for achieving the goal, even in the presence of unmodeled situations or external forces.

We studied, on a simulated\footnote{Planned onsite experiments with the real platform were not possible to finish due to university access restrictions.} 7DOF Franka Emika Panda industrial arm, 
\begin{enumerate}
    \item How multimodal encoding improves representation power and thus, state estimation and control accuracy;
    \item How the variational inference nature of our approach provides strong adaptation and robustness against unmodeled dynamics, environment parameter variations and sensory noise.
\end{enumerate}
Results show its improved performance by qualitative and statistical comparison with two baselines: proprioceptive AIF \cite{pezzato2020novel} and the Panda industrial arm built-in controller. For reproducibility, the code is publicly available at~\url{https://github.com/Cmeo97/MAIF}.

\section{AIF general formulation and notation}
\label{sec:formulation}
Here we introduce the standard equations and concepts from the AIF literature \cite{friston2010free}, and the notation used in this paper, framed for estimation and control of robotic systems~\cite{oliver2021empirical}. The aim of the robot is to infer its state (unobserved variable) by means of noisy sensory inputs (observed). For that purpose, it can refine its state using the measurements or perform actions to fit the observed world to its internal model. This is dually computed by optimizing the variational free energy, a bound on the Bayesian model evidence.

\begin{description}
\item[System variables.] State, observations, actions and their n-order time derivatives (generalized coordinates).
\begin{align}
    \Tilde{\obs} &= [\obs,\obs',\obs'',...,\obs^n] &\text{, sensors}\nonumber\\
    \Tilde{\latent} &= [\latent,\latent',\latent'',...,\latent^n] &\text{, multimodal system state}\nonumber\\
    \Tilde{\muvec} &= [\muvec,\muvec',\muvec'',...,\muvec^n] &\text{, proprioceptive state}\nonumber\\
    \Tilde{\xnoise} &= [\xnoise,\xnoise',\xnoise'',...,\xnoise^n] &\text{, sensory noise}\nonumber\\
    \Tilde{\znoise} &= [\znoise,\znoise',\znoise'',...,\znoise^n] &\text{, state fluctuations}\nonumber\\
    \action &= \{a_1,\ldots, a_m\}&\text{, action (m actuators)}\nonumber
\end{align}
Where $\obs' = d\obs/dt$. Depending on the formulation the action $\action$ can be force, torque, acceleration or velocity. In this work action refers to torque. We further define the time-derivative of the state vector $D\Tilde{\latent}$ as:
\begin{align}
    D \Tilde{\latent} = \frac{d}{dt}(\latent,\latent',\ldots,\latent^n) = [\latent',\latent'',\ldots,\latent^{n+1}]\nonumber
\end{align}

\item[Generative models.]
Two generative models govern the robot; the mapping function between the robot's state and the sensory input $\g(\Tilde{\latent})$ (e.g., forward kinematics) and the dynamics of the internal state $f(\Tilde{\latent})$.
\begin{align}
    \Tilde{\obs} &= \g(\Tilde{\latent}) + \Tilde{\xnoise} \\
    D\Tilde{\latent} &= \f(\Tilde{\latent}) + \Tilde{\znoise}
    \label{gen}
\end{align}
where $\xnoise \sim \mathcal{N}(\boldsymbol{0}, \Sigma_x )$ and $\znoise \sim \mathcal{N}(\boldsymbol{0}, \Sigma_z )$ are the sensory and process noise respectively.  $\Sigma_x$ and $\Sigma_z$ are the covariance matrices that represent the controller's confidence about each sensory input and about its dynamics respectively.
\vspace{0.2 cm}
\item[Variational Free Energy (VFE).]
The VFE is the optimization objective for both estimation and control. We use the definition of the $\mathcal{F}$ based on \cite{friston2010action}, where the action is implicit within the observation model $\obs(a)$. Using the KL-divergence the VFE is:
\begin{align}
\mathcal{F} &= \text{KL}\left[ q(\latent) || p(\latent|\obs) \right] - \log p(\obs) \\
&= \text{KL}\left[ q(\latent) || p(\latent,\obs) \right] = -\text{ELBO} \nonumber
\end{align}

\noindent \textit{State estimation} using gradient optimization:
\begin{equation}
   \boldsymbol{\dot{\Tilde{\latent}}} =  D\Tilde{\latent} - k_{z}\nabla_{\latent}\mathcal{F}({\Tilde{\obs}}, {\Tilde{\latent}})
    \label{eq:perception_general}
\end{equation}
\noindent \textit{Control} using gradient optimization:
\begin{equation}
    \dot{\action} = -k_a \sum_x \frac{d \Tilde{\obs}}{da} \cdot \nabla_{\Tilde{\obs}}\mathcal{F}({\Tilde{\obs}}, {\Tilde{\latent}})
    \label{eq:action_general}
\end{equation}
The VFE has a closed form under the Laplace and Mean-field approximations \cite{friston2007variational,oliver2021empirical,buckley2017free} and it is defined as:
\small
\begin{align}
    \mathcal{F}(\Tilde{\latent},\Tilde{\obs}) \simeq& -\ln p(\Tilde{\latent},\Tilde{\obs}) = -\ln p(\Tilde{\obs}|\Tilde{\latent}) - \ln p(\Tilde{\latent})\\
    \simeq& \;(\obs-g(\obs))^T \Sigma_x^{-1} (\obs-g(\obs)) \nonumber\\
    &+ (D\latent-f(\latent))^T \Sigma_z^{-1}(D\latent-f(\latent)) \nonumber\\
    &+ \frac{1}{2} \ln|\Sigma_x| + \frac{1}{2} \ln|\Sigma_z| \label{eq:flaplace}
\end{align}
\normalsize
The first term of Eq. (\ref{eq:flaplace}) is the sensor prediction error and the second term is the dynamics prediction error. 

\end{description}

\section{MVAE-AIF controller}
\label{sec:model}
The main novel addition presented in this work is the introduction of high-dimensional representation learning within the AIF optimization framework to provide at the same time multimodal state estimation and torque control. We first describe the architecture, the generative models and the training. Second, we detail the estimation and control equations and finally, we summarize the algorithm.

\subsection{Architecture and design}
Our proposed multimodal variational autoencoder active inference (MVAE-AIF) architecture is depicted in Fig.~\ref{fig:architecture:def}. Figures \ref{fig:architecture:visual} and \ref{fig:architecture:visual_out} shows the camera input and the predicted sensory input from the learnt generative model.

\subsubsection{Sensors}
We assume that the robot has (noisy) proprioceptive sensors that provide the joint angles $\joints$, velocities $\dot{\joints}$ and accelerations $\ddot{\joints}$. Furthermore, there is a camera that provides images $\v{I}$ of size $(w\times h)$ of the robot.

\subsubsection{Robot generative models}

The generative models of the agent are approximations of the real generative models of the world and are composed of a sensory generative function $g(\latent)$, which maps the internal state to the visual and proprioceptive spaces and a dynamics function $f(\latent)$, which describes the evolution the system.
\begin{align}
\obs_v &=   \g_{v} (\latent) + \xnoise_v &\text{\quad visual}\\
\obs_q &= \g_{q} (\latent) + \xnoise_q  &\text{\quad joints}\\
D\latent &= \f(\latent, \rhovec) + \znoise_{z}  &\text{\quad state dynamics}
\end{align}
where $\obs_v$ and $\obs_q$ are the visual and proprioceptive predictions respectively. The processes noise $\xnoise_q$, $\xnoise_v$ and $\znoise_z$ are assumed to be drawn from a multivariate normal distribution with zero mean and covariance matrices $\Sigma_{q}$,  $\Sigma_v$ and $\Sigma_z$ respectively. Note that we included $\rhovec$ into the dynamics function. This variable will describe the desired goal and acts as a prior that drives the system.

\subsubsection{Generative models and state representation} 

The defined sensory generative models are approximated by a multimodal variational autoencoder (MVAE) which maps the internal state representation to both proprioceptive and visual sensory spaces into a common latent space. We designed the MVAE using two couples of parallel encoders and decoders. The MVAE latent space describes the state of the system and the output of the decoders are the predicted image $\obs_v$ and joint values $\obs_q$. The state of the system $z$ is defined as:
\begin{equation}
    \latent = encoder_{\joints}(\joints) + encoder_{v}(\v{I})
    \label{z_eq}
\end{equation}
As a result, modalities have a low correlation, allowing sensory integration without reciprocal degradation. In other words, if one of the two modalities is not well reconstructed, the other one can still be inferred. For instance, in the case of visual occlusion, proprioceptive reconstruction is still possible.    
The robot can infer the sensory input given the state using the decoders, which act as the generative learnt models:
\begin{align}
    \obs_v = g_v(\latent) = decoder_{v}(\latent) \\
    \obs_q = g_p(\latent) = decoder_{q}(\latent) 
\end{align}

As we only need the sensory generative models, the encoders are just employed for the training. Then, in the AIF control loop, only the decoders are used. The detailed description of the network layers can be found in Table \ref{tab:network_specs}.

\noindent For algorithmic purposes we also define the proprioceptive state (joint angles belief) as $\muvec= \{g_q(\latent), \muvec', \muvec'', \muvec'''\}$, where the first component is the predicted joint angles and the others are the higher-order derivatives.

\subsubsection{Generative models learning}
The MVAE was trained with a dataset containing 500000 images with size $(128 \times 128)$ and the associated joints angle values which we created performing motor babbling on the robot arm. To accelerate the training, we included a precision mask $\Pi_{\obs_v}$, computed by the variance of all images and highlighting the pixels with more information. The augmented reconstruction loss employed was:
\begin{align}
    \mathcal{L} = \text{MSE}(\obs_v \!+\! \Pi_{\obs_v}\obs_v,\; \v{I}) + \text{MSE}(\obs_q, \joints)
\end{align}

\subsubsection{Goal definition}
We define the goal of the system following a similar approach as \cite{oliver2021empirical}. The desired goal, described in the sensory space $\rhovec=\obs_d$, is introduced in the internal dynamics as follows:
\begin{align}
f(\latent,\rhovec\!=\!\obs_d) = T(\latent) (\obs_d -g(\latent)) = \frac{\partial \g(\latent)}{\partial \latent} (\obs_d - g(\latent))
\end{align}
Where $T(\latent)$ is a function that maps the error in the sensory space to the latent space. In this case, the partial derivative of $g(\obs)$ with respect to $\latent$. The desired goal in this work is defined by the final image and joints angles $\obs_d=\{\v{I}_d, \joints_d\}$.

\subsection{Estimation and control}
The online estimation and control problem is solved by optimizing the VFE through gradient optimization, computing Equations (\ref{eq:perception_general}) and (\ref{eq:action_general}). 

\vspace{2mm}

State \textbf{estimation} uses the Laplace approximation of $\mathcal{F}$, described in Eq. (\ref{eq:flaplace}). We update the multimodal state $\latent$ by means of the partial derivative of the VFE with respect to $\latent$ ($\dot{\latent} = -k\partial_\latent\mathcal{F}$):
\small
\begin{align}
    \dot{\latent} =&  k_{v}\frac{\partial \g_{I}}{\partial z} \Sigma_v^{-1} (\obs_v - \g_{I}(\latent)) +  k_{q}\frac{\partial \g_{q}}{\partial z} \Sigma_q^{-1} (\muvec - \g_{q}(\latent)) \nonumber\\
    &+ k_{v}\frac{\partial \f_{I}}{\partial z} \Sigma_v^{-1} (\boldsymbol{I}_d - \f_{I}(\latent,\rhovec)) +  k_{q}\frac{\partial \f_{q}}{\partial z} \Sigma_q^{-1} (\joints_d - \f_{q}(\latent, \rhovec)) \nonumber\\
    \label{eq:up_z}
\end{align}
\normalsize
As we do not have access to the high order generalized coordinates of the latent space $\latent',\latent''$, we track both the multimodal shared latent space $\latent$ and the higher orders of the proprioceptive (joints) state $\muvec',\muvec''$. This is very useful as we can use the angular velocity and acceleration within the controller. Thus, we update the proprioceptive state velocity and acceleration by computing the partial derivative of $\mathcal{F}$ with respect to $\muvec',\muvec''$, as it was a unimodal proprioceptive controller~\cite{pezzato2020novel}, but the joint angles are predicted by the MVAE.
\small
\begin{align}
     \muvec =& g_q(\latent) \label{eq:update:mu}\\
     \dot{\muvec}' =& \muvec'' + k_{\mu} \left(\Sigma_{\dot{q}}^{-1} ( \dot{\joints} - \muvec') \nonumber\right.\\
     &\left.- \Sigma_{\mu}^{-1} ( \muvec' + \muvec - \joints_{d}) - \Sigma_{\dot{\mu}}^{-1} (\muvec'' + \muvec') \right) \label{eq:update:dotmu}\\
     \dot{\muvec}'' =&  - k_{\mu} \Sigma_{\dot{\mu}}^{-1} (\muvec'' + \muvec') \label{eq:update:ddotmu}
\end{align}
\normalsize
where $\Sigma_{\mu}^{-1},\Sigma_{\dot{\mu}}^{-1}$ are the precision (inverse variance) matrices related to joint angles and velocities beliefs.

\vspace{2mm}

The \textbf{action} (torque) is computed by optimizing the VFE using Eq. (\ref{eq:action_general}). Here we only consider the proprioceptive errors. In the AIF literature this models the reflex arc, where top-down proprioceptive predictions coming from higher cortical levels produce the motoric reflexes~\cite{otto2020deep}. Thus, the torque commands are updated with the following differential equation:
\small
\begin{align}
    \dot{\action} &= - k_a \left(\partial_\action \muvec \Sigma_q^{-1} (\joints - \muvec) + \partial_\action \muvec' \Sigma_{\dot{\mu}}^{-1} (\dot{\joints} - \muvec') \right) \\
    &= - k_a (\Sigma_{q}^{-1}(\joints - \muvec) + \Sigma_{\dot{\mu}}^{-1} (\dot{\joints} - \muvec'))
\end{align}
\normalsize
Although we can compute the action inverse models $\partial_\action \muvec, \partial_\action \muvec'$ through online learning using regressors~\cite{lanillos2018active}, we let the adaptive controller absorb the non-linearities of the relation between the torque and the joint velocity. Thus, as described by \cite{pezzato2020novel} we just consider the sign of the derivatives.

\subsection{Algorithm}
Algorithm \ref{alg:mvae-aic} describes the proposed controller. The first stage solves the free energy optimization using the prediction errors between the generative model and the sensor measurements. We compute $\dot{z}_{q}, \dot{z}_{v}, \dot{z}_{q_d}$ and $\dot{z}_{v_d}$ as the four terms in Eq. \ref{eq:up_z}. Since these terms are equivalent to backpropagation operations over the respective errors, as we proposed in \cite{sancaktar2020end}, they are computed by defining a node with $\dot{z}$ on the network virtual tree and backpropagating errors through the decoders. Once we have the contributions from the sensory and dynamics prediction errors, we update the robot multimodal state and the proprioceptive higher-order derivatives. Finally, we compute the action to minimize the inverse variance weighted proprioceptive prediction errors.

\begin{algorithm}[hbtp!]
\caption{MVAE-AIF}
\label{alg:mvae-aic}
\begin{algorithmic}
\Require $\obs_{desired}=\{\v{I}_d,\joints_d\}$
\While {$\neg goal$}
\vspace{0.1 cm}
\State $\joints,\v{I} \gets$ Sensors (joint values, camera)
\vspace{0.2 cm}
\State $Compute \hspace{0.1 cm} prediction \hspace{0.1 cm} errors \hspace{0.1 cm} in \hspace{0.1 cm} \latent \hspace{0.1 cm} space, \hspace{0.1 cm}Eq. \hspace{0.1 cm}\ref{eq:up_z}$
\State $\obs_{v} = \g_{v}(\latent)   \quad\quad Decoder \hspace{0.1 cm} visual $
\State $\obs_{q} = \g_{q}(\latent)  \quad\quad Decoder \hspace{0.1 cm} joints$
\State $\dot{\latent}_{q} = \g_{q}.backward(\Sigma^{-1}_{q}(\joints- \obs_q))$
\State $\dot{\latent}_{v} = \g_{v}.backward(\Sigma^{-1}_{v}(\v{I} - \obs_v))$
\State $\dot{\latent}_{q_d} = \g_{q}.backward(\Sigma^{-1}_{q}(\joints_{d} - \obs_q))$
\State $\dot{\latent}_{v_d} = \g_{v}.backward(\Sigma^{-1}_{v}(\v{I}_{d} - \obs_v))$
\vspace{0.2cm} 
\State $Update \hspace{0.1 cm} state $
\State $\dot{\latent} = z_q+z_v+z_{q_d}+z_{v_d}$
\State $\dot{\muvec}', \dot{\muvec}''  \gets  proprioceptive \hspace{0.1 cm} state \hspace{0.1 cm} Eq. \hspace{0.1 cm} \ref{eq:update:dotmu}, \hspace{0.1 cm}  \ref{eq:update:ddotmu}$
\vspace{0.2 cm}
\State $Active \hspace{0.1 cm} inference$
\State $\dot{\action} = - k_a \Sigma_q^{-1}(\joints - \obs_q) + \Sigma_{\dot{q}}^{-1} (\dot{\joints} - \muvec') $
\vspace{0.2 cm}
\State $Euler \hspace{0.1 cm} integration$
\State $\left[\latent, \muvec', \muvec''\right] \mathrel{+}= \delta_t \left[\dot{\latent},\dot{\muvec}', \dot{\muvec}''\right]$
\EndWhile 
\end{algorithmic}
\end{algorithm}

\section{Results}
\label{sec:results}
\subsection{Experiments and evaluation measures}
We performed three different experimental analyses and compared our approach (MAIF) against two state-of-the-art controllers: a proprioceptive torque AIF controller (PAIF)~\cite{pezzato2020novel} and the built-in Panda controller (BPC) with the factory parameters. Video footage of the experiments can be found in the supplementary video: \url{https://github.com/Cmeo97/MAIF} 
\begin{enumerate}
    \item \textbf{Qualitative analysis} in sequential reaching (Sec. \ref{sec:results:qualitative}). We evaluated the algorithm accuracy and behavior using several metrics, as well as studied the effect of including the visual modality. Precisely, we analyzed how the multimodal representation power affects accuracy. Besides, we show how our algorithm can work in an imaginary regime by mentally simulating the behavior.
    \item \textbf{Statistical evaluation} in 1000 random goals (Sec. \ref{sec:results:stat}). We compared our proposed controller's average accuracy and robustness to sensory noise against the baselines PAIF and BPC.
    \item \textbf{Adaptation study} (Sec. \ref{sec:results:adapt}). We evaluated the response of the system to unmodeled dynamics and environment variations by altering: the world gravity (Jupiter experiment), the compliance of the robot, and the sensory noise. Here we also compared our approach with the PAIF and the BPC.
\end{enumerate}

In the experiments, we used the following evaluation metrics.
\begin{itemize}
    \item  \textit{Joints perception error}. Error between the inferred (belief) and the observed joint angle. The more accurate are the predictions, the lower will be the perception error. 
    \item  \textit{Joints goal error}. Error between the current joint angles and the desired ones (goal).
    \item \textit{Joints belief-goal error}. Error between the inferred joint angle and the desired joints goal. It shows how the perceived joints approach the final goal.
    \item \textit{Image reconstruction error}. Error between the predicted visual input and the observed image. It describes the goodness of the visual generative model. 
    \item \textit{End-effector error}. Error between the current end-effector location and the desired one (goal). It provides a single metric to evaluate the precision of the controller.
\end{itemize}

Summarizing, joints perception and image reconstruction errors measure how well the state is estimated, while joints goal, joints belief-goal and end-effector errors give a measure of how well the control task is executed.

\subsection{Experimental setup and parameters}
\label{sec:results:experimental_setup}

 Experiments were performed on the Gazebo simulator and the 7DOF Franka Panda robot arm~\cite{franka} using ROS~\cite{koubaa2019ROS} as interface and Pytorch~\cite{stevens2020pytorch} for the MVAE. A camera model was used to acquire visual grey scaled images with size 1x128x128 pixels. The camera was placed in front of the robot arm with a distance of $1.2$ m. The average frequency of the AIF algorithm was $110$ Hz. Proprioceptive and Visual sensors have publishing frequencies of $1000$ Hz and $10$ Hz respectively. Due to this discrepancy, they were synchronized updating them once per AIF loop iteration.
 
 The tuning parameters for the AIF controller are:
\begin{itemize}
    \item $\Sigma_{q}, \Sigma_{v}$: Variances representing the confidence about sensory data were set as the variances of the training dataset (joints and images).
    
    \item $\Sigma_{\mu}\!\!=\!\!2.5, \Sigma_{\mu'}\!\!=\!\!1.0$: Variances representing the confidence of internal belief about the states;
    \item $k_{\mu}\!\!=\!\!11.67,k_q\!\!=\!\!0.6,k_v\!\!=\!\!0.1,k_a\!\!=\!\!900$:  The learning rates for state update and control actions respectively were manually tuned in the ideal settings experiment. 
\end{itemize}

Table \ref{tab:network_specs} details the specification of the encoders and decoders used in the MVAE. For training the MVAE we set the number of training epochs to 100, and we used ADAM~\cite{kingma2017adam} as the optimization algorithm. All experiments are simulated on a computer with CPU: AMD ryzen 9 3900X, GPU: Nvidia GeForce RTX 3080 msi. \newpage

\hspace{0.1 cm}
\vspace{0.05 cm}
\hspace{0.1 cm}

\begin{table}[hbtp!]
\scalebox{0.93}{\begin{tabular}{|l|l|l|l|}
\hline
\textbf{Visual Enc.} & \textbf{Prop Enc.} & \textbf{Visual Dec.} & \textbf{Prop Dec.} \\ \hline
\textit{Input} - $\v{I}$   & \textit{Input} - $\joints$    & \textit{Input} - $\latent$  & \textit{Input} - $\latent$ \\ 
Conv(1, 128, 3)   & FC(7, 64)        & upConv(1, 16, 4)  & FC(256, 128) \\ 
Conv(128, 64, 3)  & FC(64, 512)      & upConv(16, 32, 4) & FC(128, 64)  \\ 
Conv(64, 32, 3)   & FC(512, 4096)    & upConv(32, 16, 4) & FC(64, 7)    \\ 
upConv(32, 32, 3) & Conv(16, 1, 2)   & upConv(16, 1, 4)  & \textit{Output} - $\obs_p$\\ 
Conv(32 ,16, 2)   & \textit{Output} - $\latent$ & \textit{Output} - $\obs_v$  &              \\ 
upConv(16, 16, 4) &                &                   &              \\ 
Conv(16, 1, 2)    &                &                   &              \\ 
\textit{Output} - $\latent$    &                &                   &              \\ \hline
\end{tabular}}
\caption{Network layers specifications. Type: Convolution (Conv), Trasposed Convolution (upConv), Fully connected (FC). All layers use ReLu activation function, except for the visual decoder output that uses $\tanh(\text{ReLu})$. MaxPool and AvgPool are used to increase network generalization.}
\label{tab:network_specs}
\end{table} 

\subsection{Qualitative analysis in a sequential reaching task}
\label{sec:results:qualitative}
\begin{figure}[hbtp!]
    \centering
	\subfloat[Home]{
		\centering
		\includegraphics[width=0.21\columnwidth]{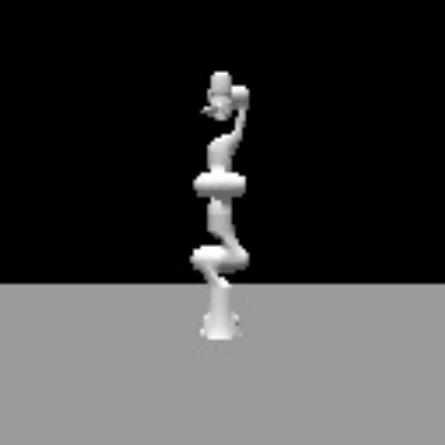}
		\label{fig:goal_pose:1}
	}
	\subfloat[$\v{I}_{d_1}$]{
		\centering
		\includegraphics[width=0.21\columnwidth]{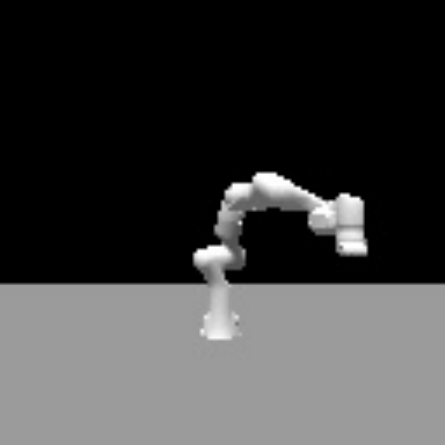}
		\label{fig:goal_pose:2}
	}
	\subfloat[$\v{I}_{d_2}$]{
		\centering
		\includegraphics[width=0.21\columnwidth]{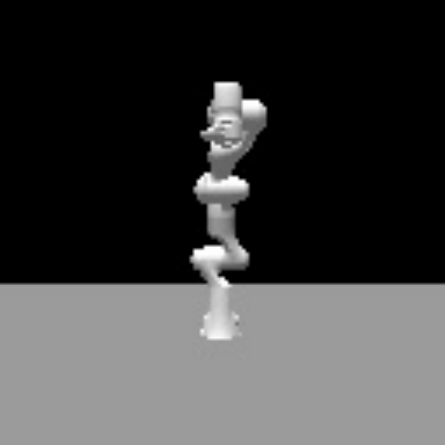}
		\label{fig:goal_pose:3}
	}
	\subfloat[$\v{I}_{d_3}$]{
		\centering
		\includegraphics[width=0.21\columnwidth]{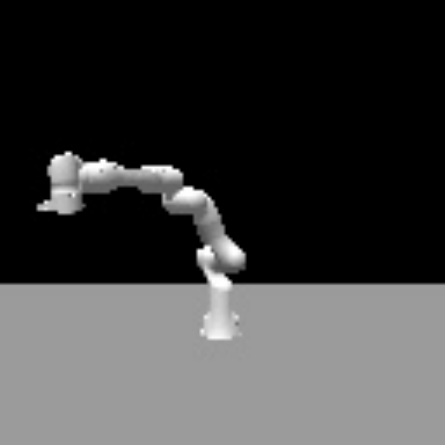}
		\label{fig:goal_pose:4}
	}
	\caption{Robot home position and visual goals.}
	\label{fig:goal_pose}
\end{figure}

To analyse the controller behavior, we designed a sequential reaching task with five desired goals defined by the final joint angles $\{\joints_{d_1}, \joints_{d_2}, \joints_{d_3}, \joints_{d_4}, \joints_{d_5}\}$:
\begin{itemize}
    \item $\boldsymbol{q_{d_1}} = \left[\hspace{0.27 cm} 1.0,\hspace{0.1 cm} 0.5,\hspace{0.1 cm} 0.0,\hspace{0.05 cm} -2.0,\hspace{0.1 cm} 0.0,\hspace{0.1 cm} 2.5,\hspace{0.1 cm} 0.0 \hspace{0.1 cm}  \right]$ rad
    \item $\boldsymbol{q_{d_2}} = \left[\hspace{0.27 cm} 0.0,\hspace{0.1 cm} 0.2,\hspace{0.1 cm} 0.0,\hspace{0.05 cm} -1.0,\hspace{0.1 cm} 0.0,\hspace{0.1 cm} 1.2,\hspace{0.1 cm} 0.9 \hspace{0.1 cm} \right]$ rad
    \item $\boldsymbol{q_{d_3}} = \left[-1.0,\hspace{0.1 cm} 0.5,\hspace{0.1 cm} 0.0,\hspace{0.05 cm} -1.2,\hspace{0.1 cm} 0.0,\hspace{0.1 cm} 1.6,\hspace{0.1 cm} 0.0 \hspace{0.1 cm} \right]$ rad
    \item $\boldsymbol{q_{d_3}} = \boldsymbol{q_{d_2}}$
    \item $\boldsymbol{q_{d_5}} = \boldsymbol{q_{d_1}}$
\end{itemize}
and the desired visual input $\{\v{I}_{d_1},\v{I}_{d_2},\v{I}_{d_3},\v{I}_{d_4}\!\!=\!\!\v{I}_{d_2}, \v{I}_{d_5}\!\!=\!\!\v{I}_{d_1}\}$, depicted in Fig.~\ref{fig:goal_pose}. In all experiments the robot starts in home position ($\joints_{home} \!\!=\!\! \v{0}$ rad), corresponding to the arm straight up.

\subsubsection{Sequential reaching}
\begin{figure}[hbtp!]
	\subfloat[Joints goal error]{
		\centering
		\includegraphics[scale=0.26]{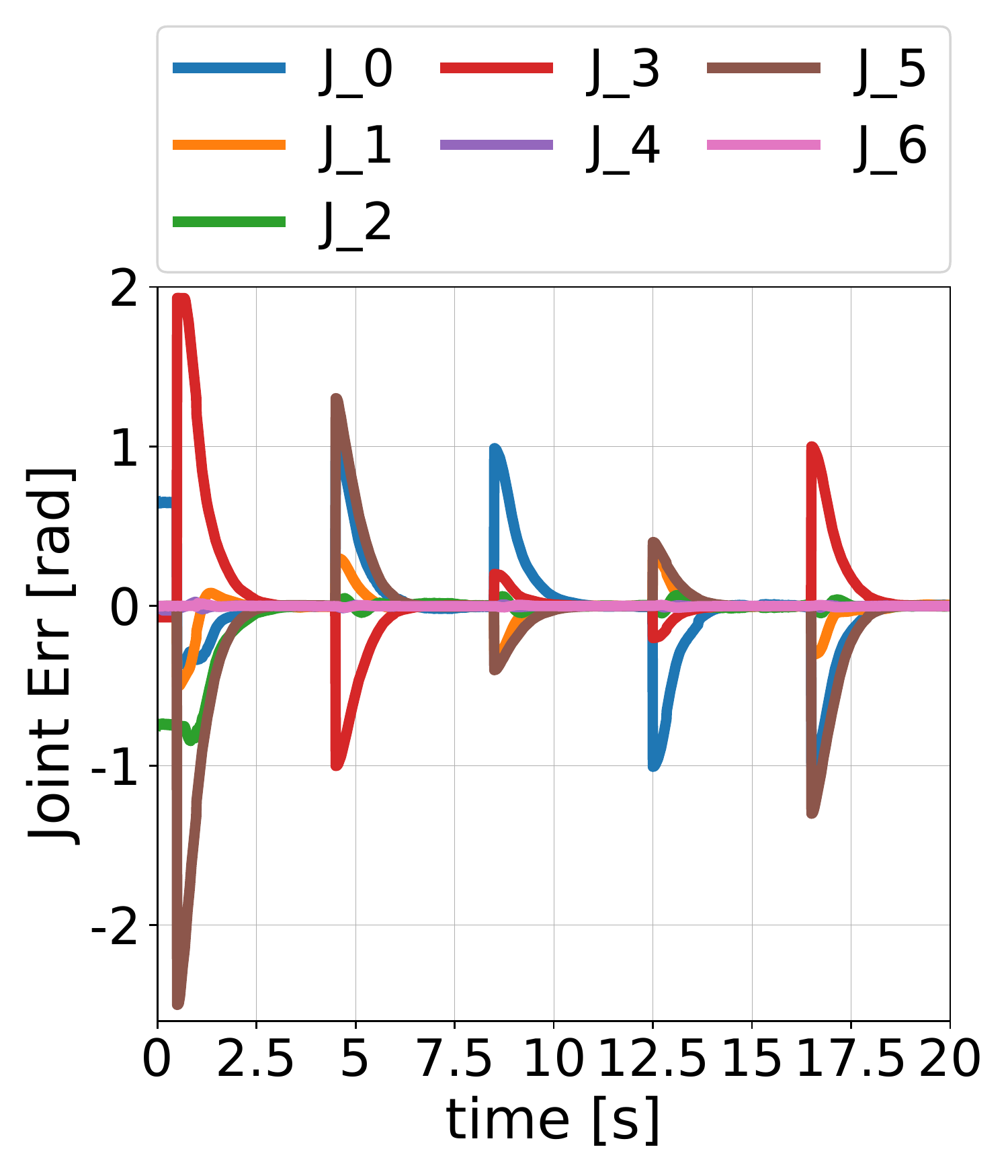}
		\label{fig:results:1}
	}
	\hspace{0.01 cm}
	\subfloat[Image reconstruction error]{
		\centering
		\includegraphics[scale=0.25]{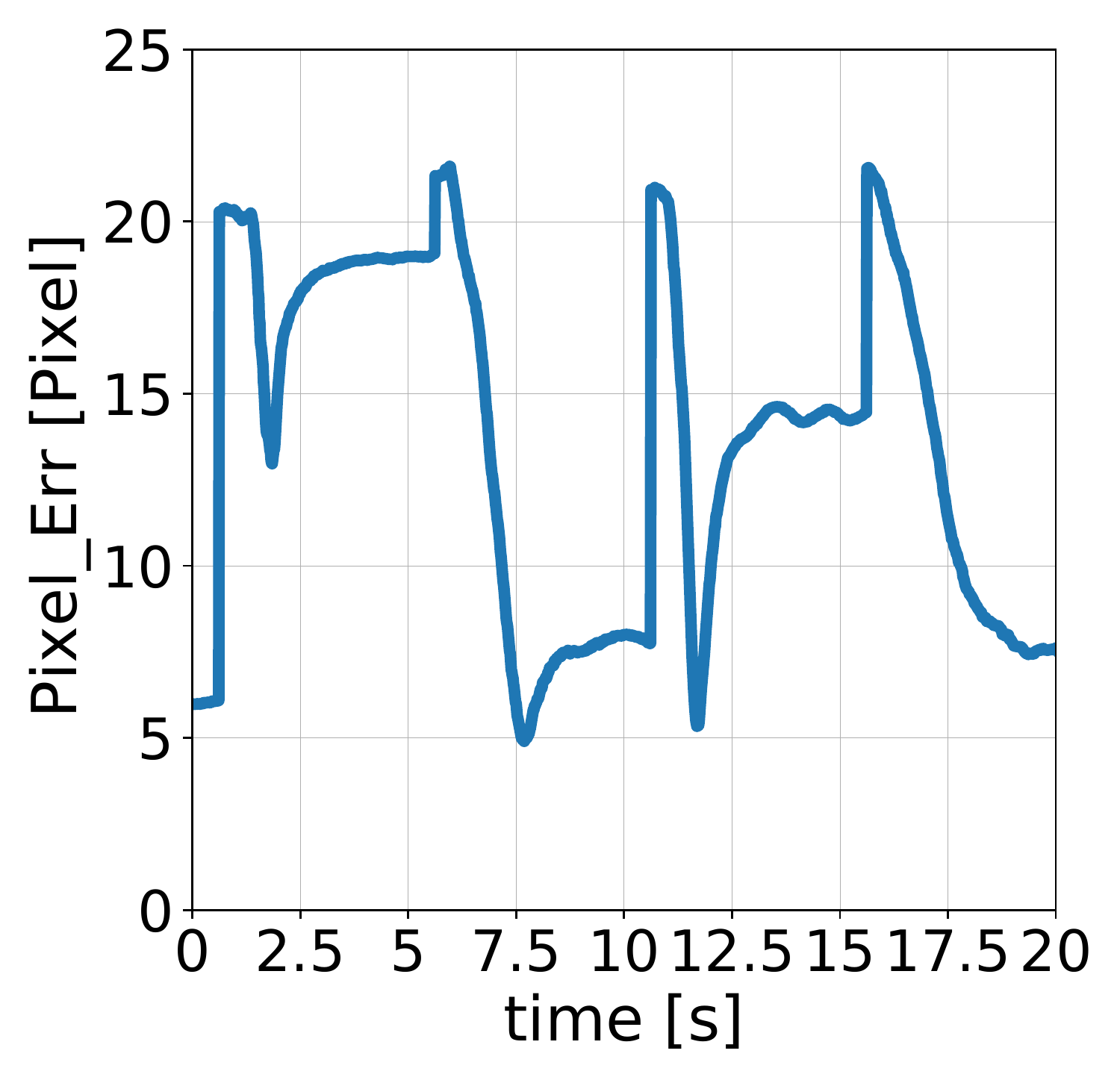}
		\label{fig:results:2}
	}\\
	\subfloat[Joints perception error]{
		\centering
		\includegraphics[scale=0.26]{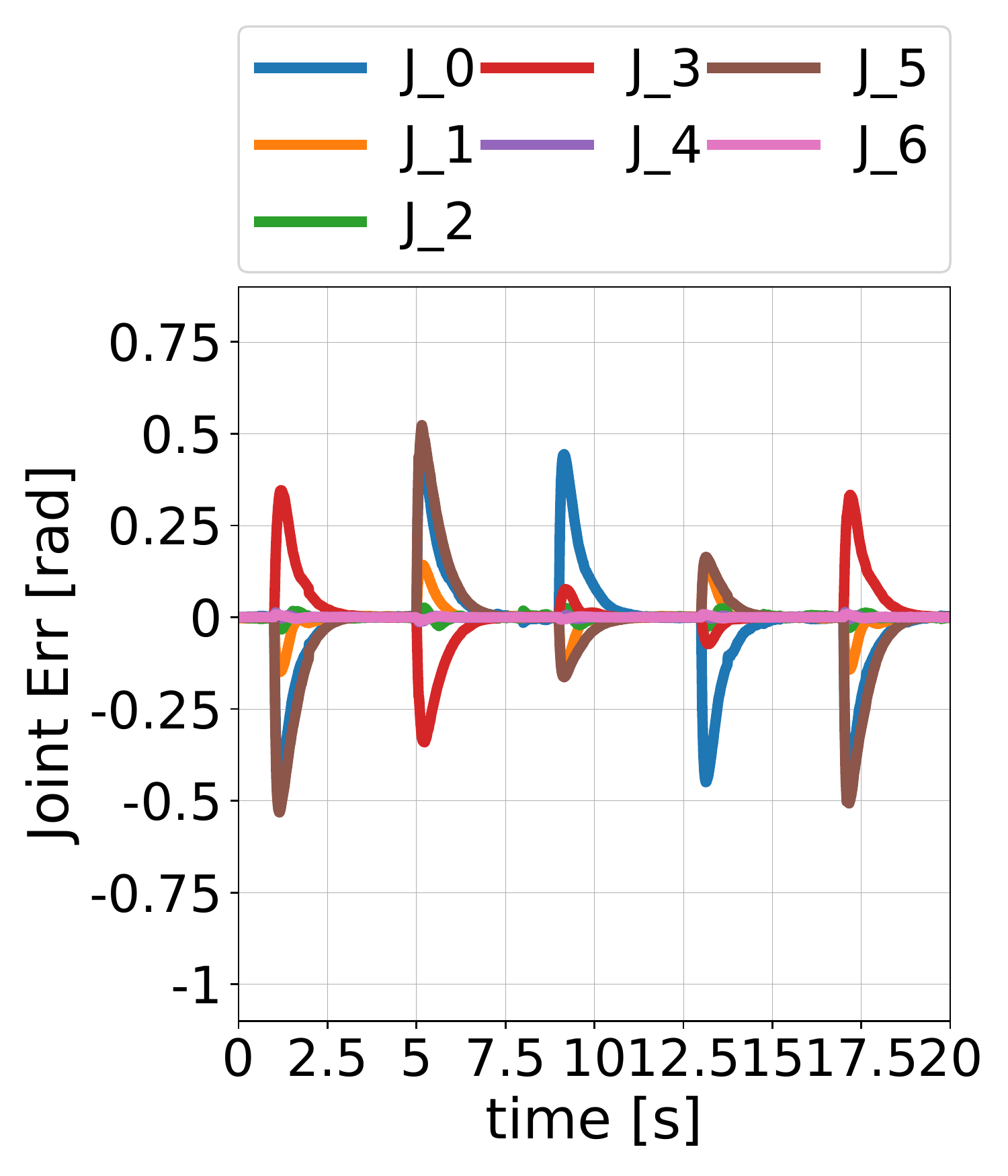}
		\label{fig:results:3}
	}
	\hspace{0.01 cm}
	\subfloat[Joints belief-goal errors]{
		\centering
		\includegraphics[scale=0.24]{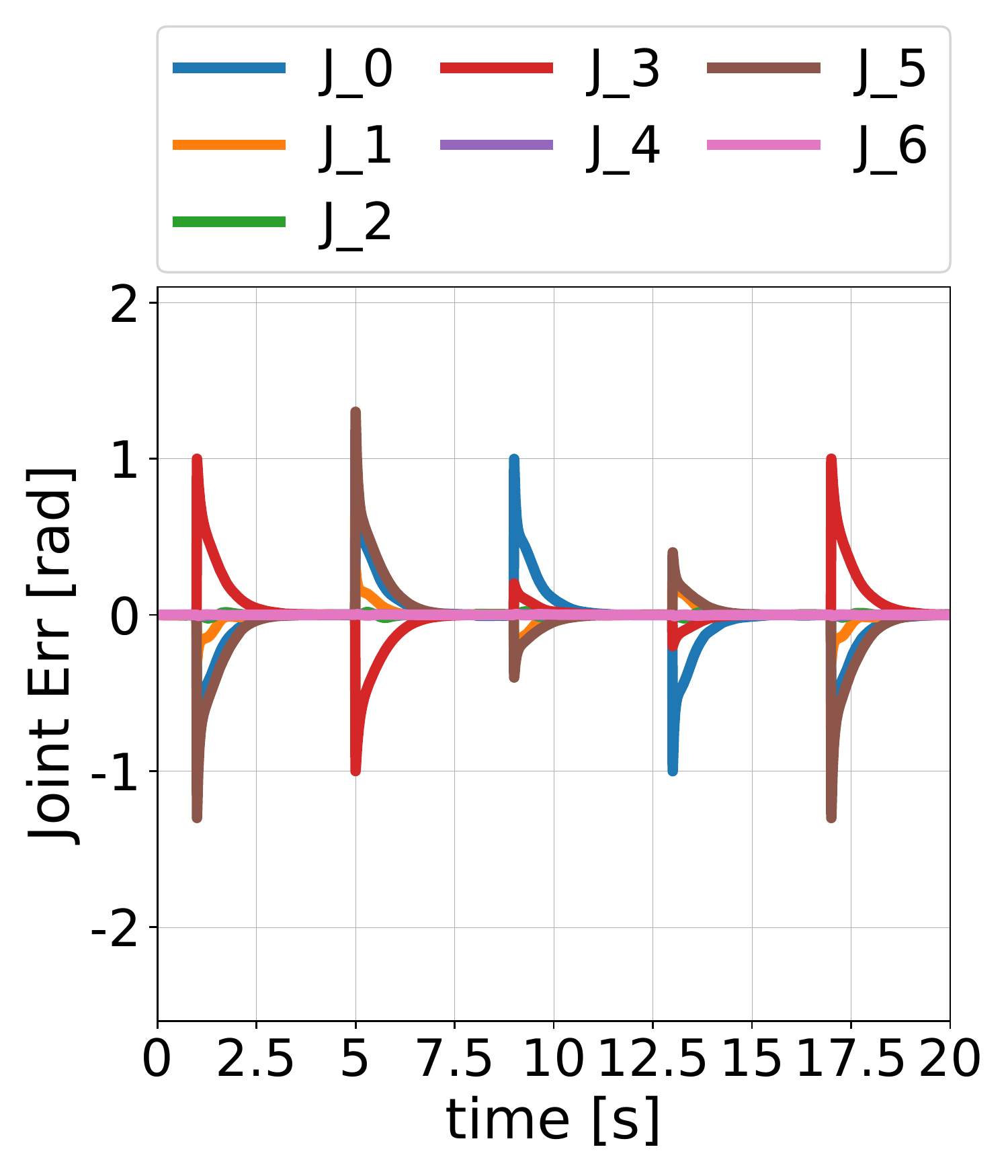}
		\label{fig:results:4}
	}\\
	\subfloat[Sequence of the MVAE predicted visual input $\g_v(\latent)$.]{
		\centering
		\includegraphics[width=0.48\textwidth]{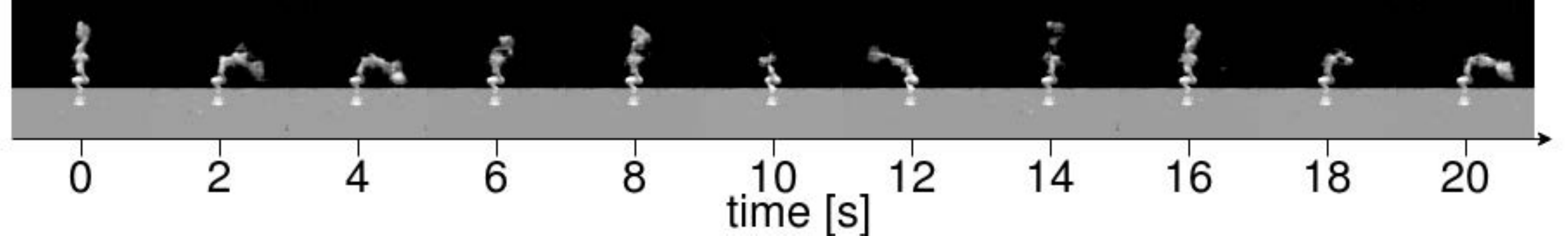}
		\label{fig:results:5}
	}
	\caption{Qualitative analysis of the error measures in the sequential reaching of five goals. All errors present peaks when the new goal is set. (a) Each line represents the error between the i-th joint angle and the desired one. (b) Image reconstruction error. (c) Each line represents the error between the i-th joint belief and the ground truth. (d) Each line represents the error between the i-th joint belief and the desired joint angle. (e) Sequence of the predicted images by the generative model along the trajectory.}
	\label{fig:results}
\end{figure}

\begin{figure}[hbtp!]
\centering
    \includegraphics[width=0.7\columnwidth, height=180px]{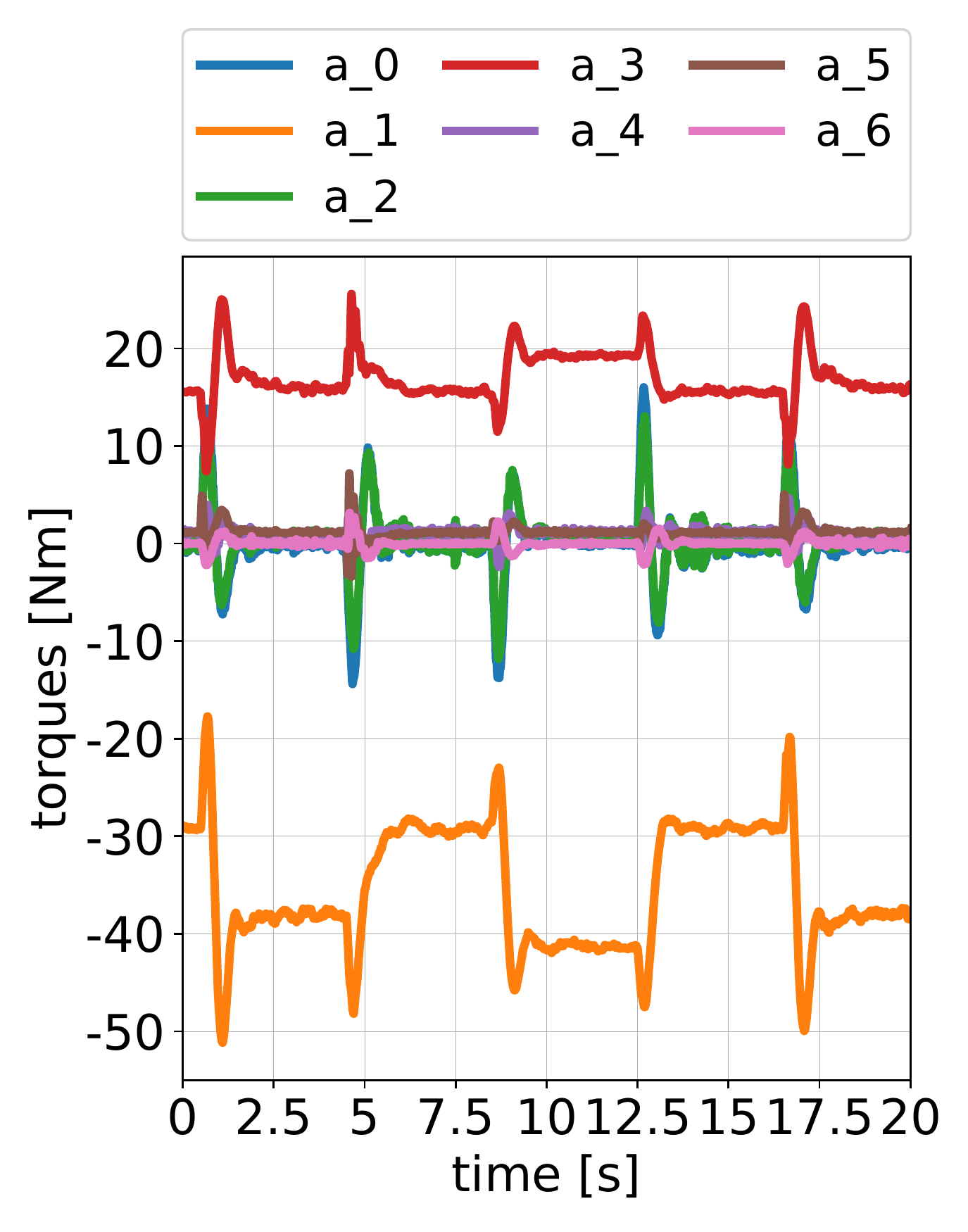}
    \caption{Torque commands for the 7-DOF robot arm using our MAIF in sequential reaching of five goals.}
    \label{fig:torques}
\end{figure}

Fig. \ref{fig:results} describes the evolution of the evaluation metrics. The joint errors (Fig. \ref{fig:results:1}) show synchronous convergence without overshooting or steady-state errors. Fig. \ref{fig:results:3} and \ref{fig:results:4} show perception and joints belief error respectively. The delay between the convergence of joints beliefs and the real ones is due to the AIF framework. The robot updates its internal belief by approximating the conditional density, maximizing the likelihood of the observed sensations and then it generates an action that results in a new sensory state, which is consistent with the current internal representation. Fig. \ref{fig:results:2} shows the visual generative model image reconstruction errors through the simulation. From one goal to the next one the error drops down. However, some goals can be better reconstructed than others, resulting in different steady errors. The reason is that many $\latent$ solutions lead to similar images. Fig.  \ref{fig:results:5} shows some visual predictions in the 20 seconds simulation. Finally, the generated actions (torques) using MAIF, for the 7 joints during the experiment, are showed in Fig. \ref{fig:torques}. Switching to a new goal generated torques that minimize prediction errors.

\subsubsection{Visual modality study}
We evaluated the effect of adding the visual modality. To this end, we tested the algorithm in the presence of visual noise and in the absence of visual input (camera occlusion or broken) by clamping the image to zeros. We compared our algorithm with the proprioceptive-AIF (PAIF) baseline. Visual noise was implemented as additive noise sampled from a Normal distribution $\xnoise_{I} \sim \mathcal{N}(\boldsymbol{0}, \Sigma_I = 0.25 )$. 

\begin{figure}[hbtp!]
\centering
    \includegraphics[width=0.8\columnwidth]{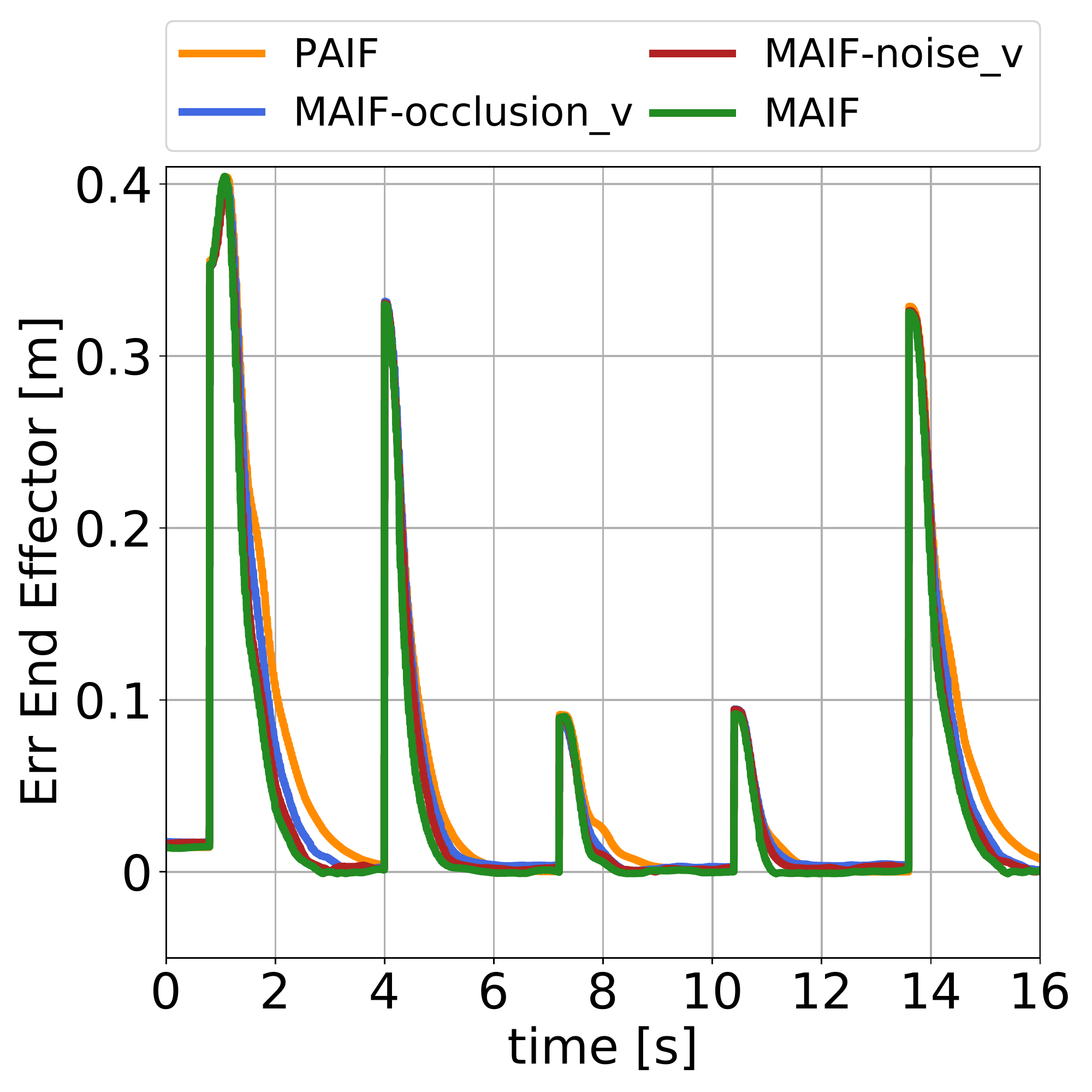}
    \caption{Visual influence in sequential reaching in ideal conditions. Lines describe the end-effector location RMSE to goal. Peaks are present when the new goal is set.}
    \label{fig:err_end_effector}
\end{figure}

Fig. \ref{fig:err_end_effector} shows the end-effector location errors using four different algorithm settings: MAIF, MAIF with visual noise, MAIF with visual occlusion and PAIF. MAIF had the best response while PAIF scored the lowest. In the case of visual occlusion, the MAIF approached the PAIF response, and in the case of noisy visual input, accuracy fell in the middle of MAIF and MAIF with visual occlusion.

\subsubsection{Mental simulation}
Unlike most of the controllers present in literature, a great advantage of using our approach is the possibility to perform simulations on the robot's brain exploiting the generative nature of the MVAE. In other words, given $\obs_{desired}$, the entire experiment can be mentally simulated. Since sensory data are not available, the state update law becomes:
\begin{equation}
    \dot{\latent} = \dot{z}_{q_d}+\dot{z}_{v_d}
\end{equation}
As a result, decoding the new internal state, the updated $\{\v{I},\joints\}$ can be computed and the new errors can be back-propagated again, creating a loop that allows the system to do imaginary simulations. 

\begin{figure}[hbtp!]
	\centering
	\subfloat[Joints errors]{
		\centering
		\includegraphics[width=0.45\columnwidth]{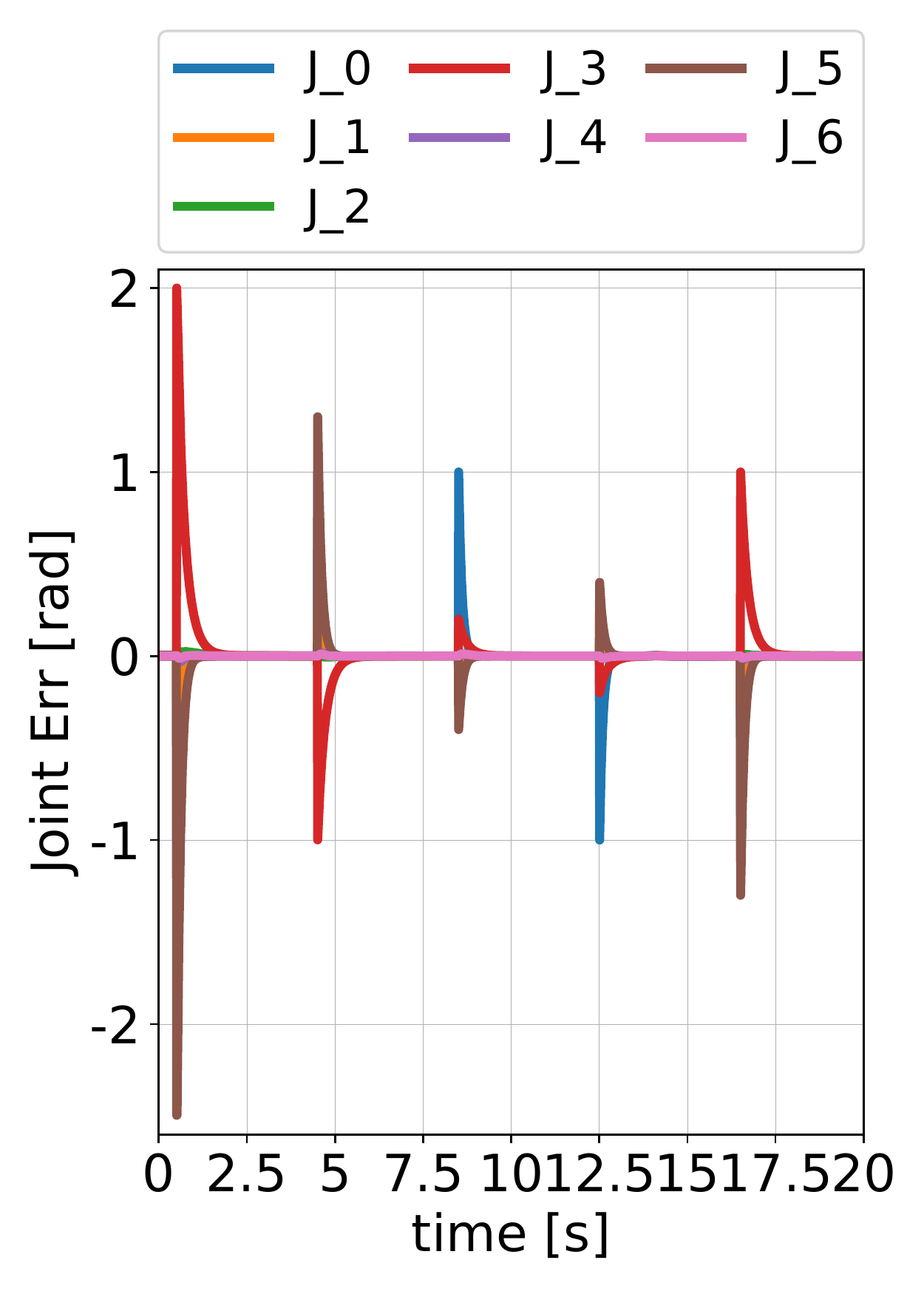}
		\label{fig:im_results:1}
	}
	\hspace{0.01 cm}
	\subfloat[Image reconstruction error]{
		\centering
		\includegraphics[width=0.45\columnwidth]{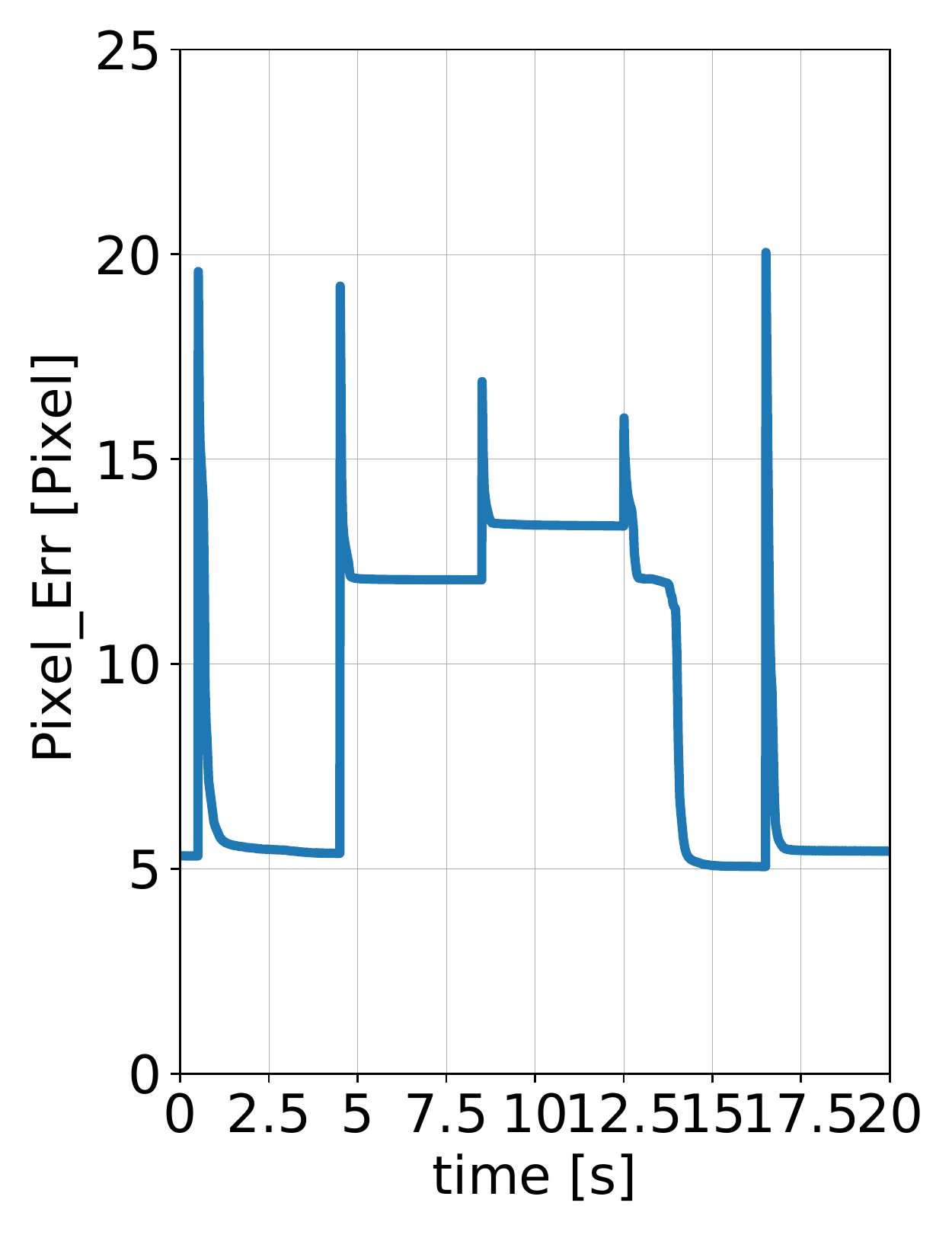}
		\label{fig:im_results:2}
	}
	\caption{Mental simulation of sequential reaching of five goals. The goal is updated on time steps where peaks are present. (a) Joints errors of an imagined simulation. Each line represents the error of the i-th joint. (b) Image reconstruction errors of an imagined simulation.}
	\label{fig:imagined_results}
\end{figure} 

Fig. \ref{fig:im_results:1} and \ref{fig:im_results:2} show respectively imagined joints error and images reconstruction error through the entire simulation. These results show that the errors converge faster to zero than in the normal regime (Fig. \ref{fig:results:4}) as it does not need to accommodate the real dynamics of the robot.

\subsection{Statistical analysis}
\label{sec:results:stat}
A statistical comparison between our MAIF, PAIF and BPC was performed to evaluate the controllers' performances on reaching tasks under normal and noisy conditions. To this end, we created a list of $1000$ random desired goals that we executed in each controller. Then, we compared their average joint errors.
\begin{figure}[hbtp!]
	\subfloat[No noise]{
		\centering
		\includegraphics[width=0.44\columnwidth]{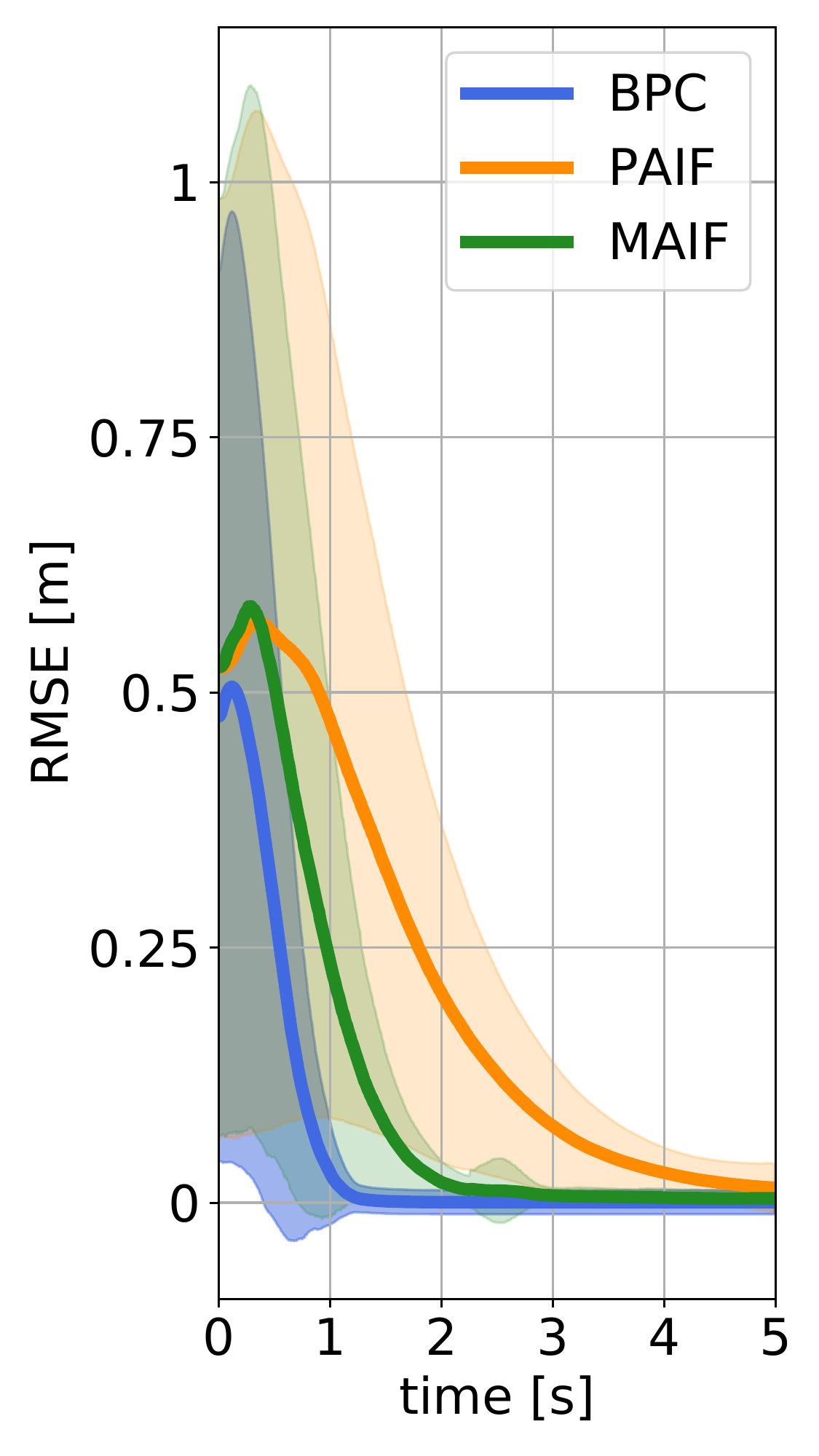}
		\label{fig:stat_results:1}
	}
	\hspace{0.01 cm}
	\subfloat[Proprioceptive noise]{
		\centering
		\includegraphics[width=0.44\columnwidth]{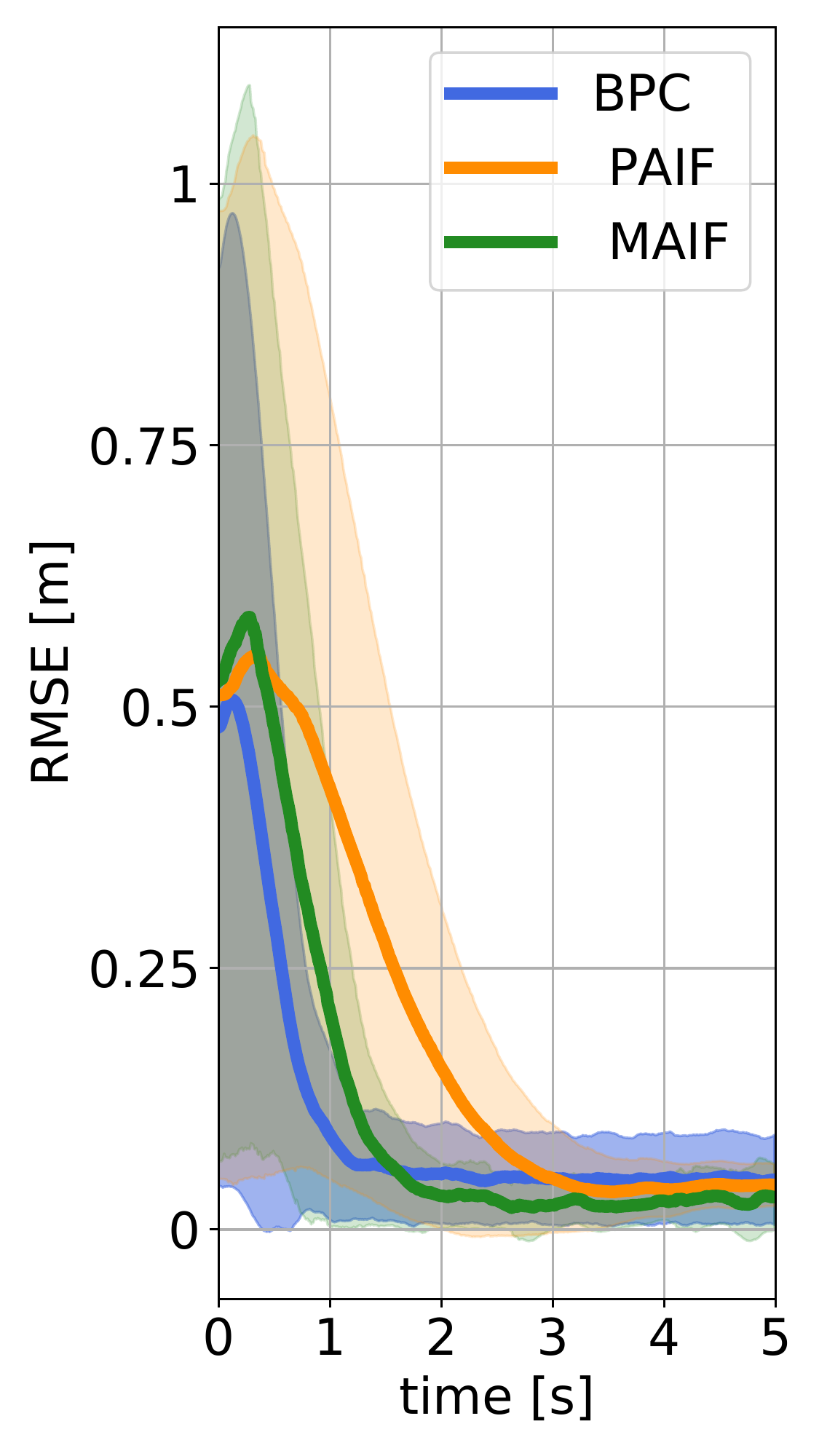}
		\label{fig:stat_results:2}
	}
	\caption{Statistical comparison of our MAIF (green), PAIF (orange) and the BPC (blue) over 1000 executions with different random goals in ideal conditions (a) and with proprioceptive noise (b). Lines represent the Root Mean Square Error~(RMSE) of the end-effector and shadings describe their standard deviation.}
	\label{fig:stat_results}
\end{figure} 
In every execution, the robot started from the home position. Fig. \ref{fig:stat_results} shows the root-mean-square error (RMSE) between the end-effector position and the desired pose over all executions for the three controllers.

In perfect conditions (Fig.\ref{fig:stat_results:1}), BPC was the fastest and had the lowest RMSE on average. MAIF had high accuracy and presented a small overshoot when reaching the goal. PAIF was much slower and had lower accuracy. However, when injecting Gaussian noise ($\xnoise_q \sim \mathcal{N}(\boldsymbol{0}, \Sigma_q = 0.5 )$) in the joint sensor measurements (Fig. \ref{fig:stat_results:2}), our proposed MAIF scored the lowest RMSE and standard deviation. Hence, MAIF presented higher robustness to proprioceptive noise with more stable behavior than the BPC and the PAIF. BPC had the highest standard deviation, showing poor robustness to sensory noise, especially we approaching the desired goal. By visual inspection of the robot behavior, we observed strong oscillations in the BPC. The robustness of MAIF is due to the algorithm structure. The free energy optimization scheme and the decoder interpolation make that the sensory noise has a low effect on the state estimation.

\subsection{Adaptation}
\label{sec:results:adapt}
To investigate our approach adaptability to unmodeled dynamics and environment variations we tested the controllers in three experiments. First, we changed the gravity of the world as if the robot was in Jupiter. Second, we altered the motors stiffness parameter of the Panda robot. Third, we reevaluated the controller in the presence of sensory noise focusing on the robot behavior. Again, we compared our algorithm (MAIF) with the built-in Panda controller (BPC) and the proprioceptive controller (PAIF). All controllers parameters were the same as in the previous experiments. Thus, no retuning was done.

\subsubsection{Jupiter experiment}
This experiment aims to show the controller performance in case of gravity changes. The reasoning behind this experiment is to evaluate MAIF performance under unexpected external conditions, such as external forces.
\begin{figure}[hbtp!]
    \centering
    \includegraphics[width=0.8\columnwidth]{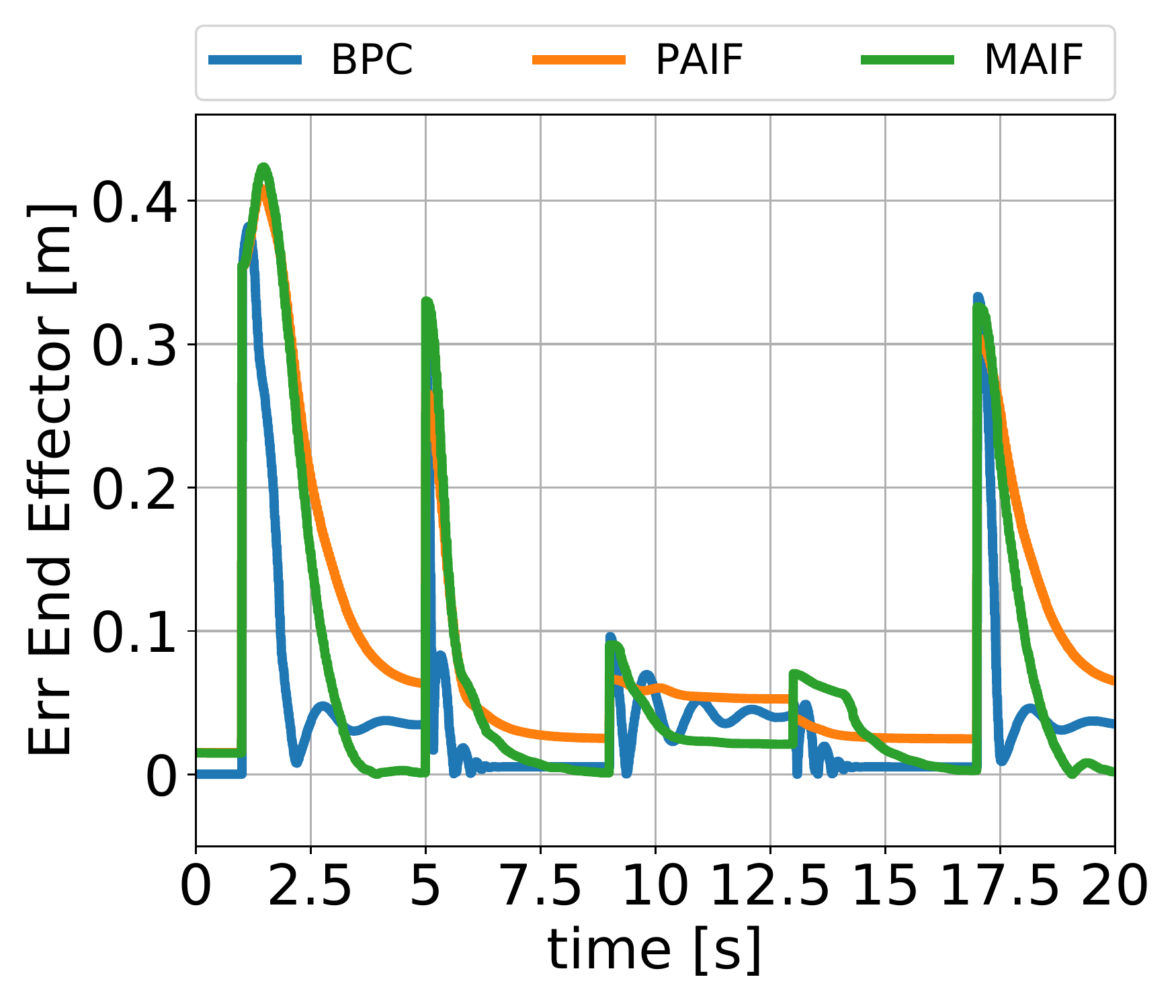}
    \caption{Jupiter experiment. The lines describe the end-effector error to the desired goal under Jupiter gravity in a reaching task with five sequential goals. MAIF (green), PAIF (orange) and BPC (blue). All end-effector errors present peaks when the new goal is set.}
    \label{fig:jupiter_exp}
\end{figure}
To do that, we performed the simulation with Jupiter gravity, which is $g = 24.79 \frac{m}{s^2}$ on the planet's surface. Fig. \ref{fig:jupiter_exp} illustrates joint angle absolute errors for each controller. MAIF was the least affected by the change of gravity. Under BPC the robot arm oscillated around the desired pose and had steady-state errors. PAIF response was also affected by the gravity change. MAIF registered a faster response than PAIF due to multisensory integration: images are not affected by gravity.

\subsubsection{Compliant experiment}
\begin{figure}[hbtp!]
    \centering
    \includegraphics[width=0.8\columnwidth]{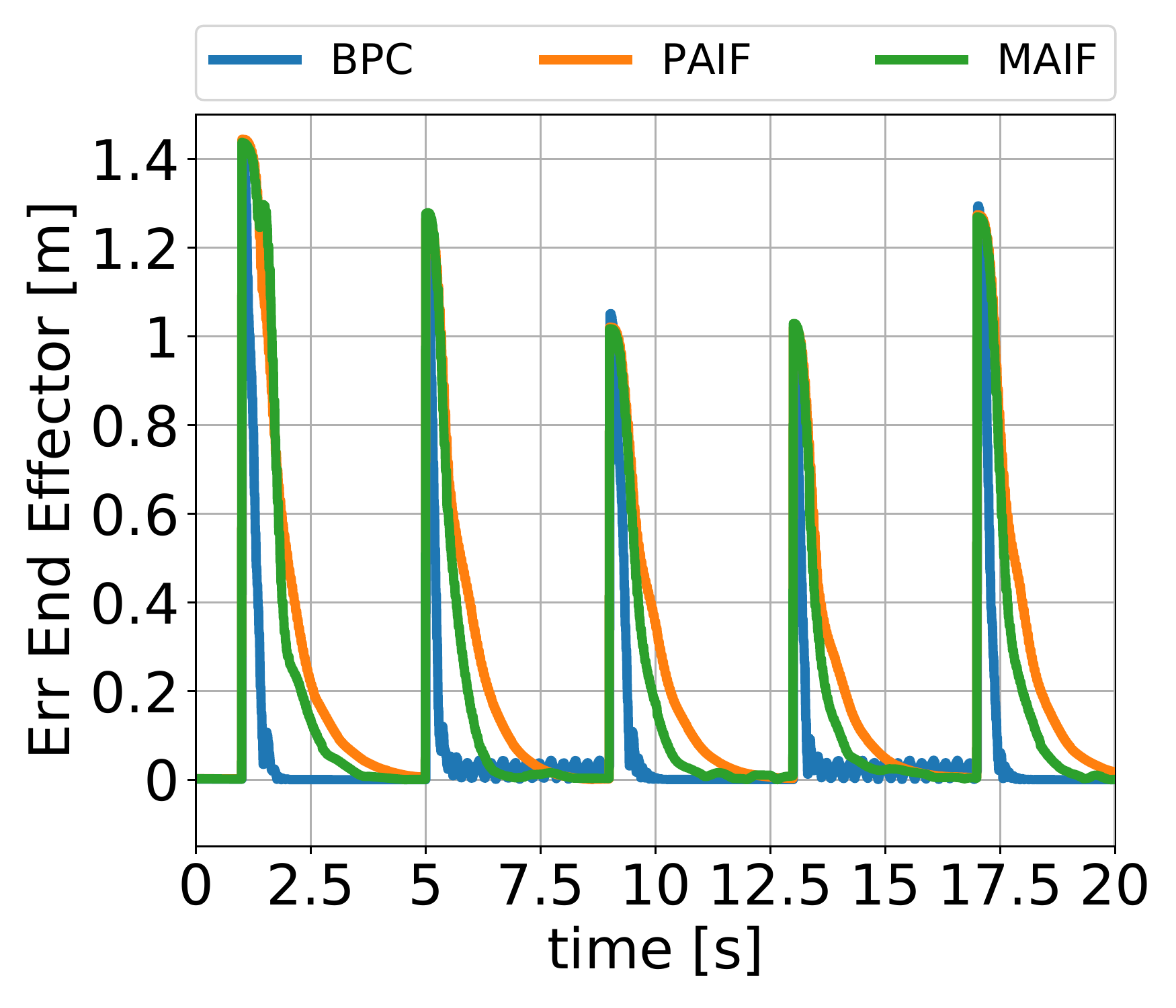}
    \caption{Compliant experiment. Sequential reaching of five goals when motors stiffness is changed from $0.61 \frac{Nm}{rad}$ to $0.01\frac{Nm}{rad}$.  All end-effector location absolute errors present peaks when the new goal is set. }
    \label{fig:compliant_exp}
\end{figure}

The goal of this experiment is the evaluation of MAIF adaptation under changes on the robot's physical parameters. We altered the robot arm's compliance by changing the motors stiffness values from $0.61 \frac{Nm}{rad}$ to $0.01\frac{Nm}{rad}$. Figure \ref{fig:compliant_exp} shows the absolute joints errors for each controller. As in the previous experiment, the BPC makes the robot arm oscillate around some goal poses, while both MAIF and PAIF do not. With low stiffness, MAIF and PAIF slightly overshot when the desired goal was far away. However, since MAIF updates his state from both proprioceptive and visual sensory data, it had much lower overshooting and faster response than PAIF.

\subsubsection{Sensory noise experiment}
\begin{figure}[hbtp!]
    \centering
    \includegraphics[width=0.8\columnwidth]{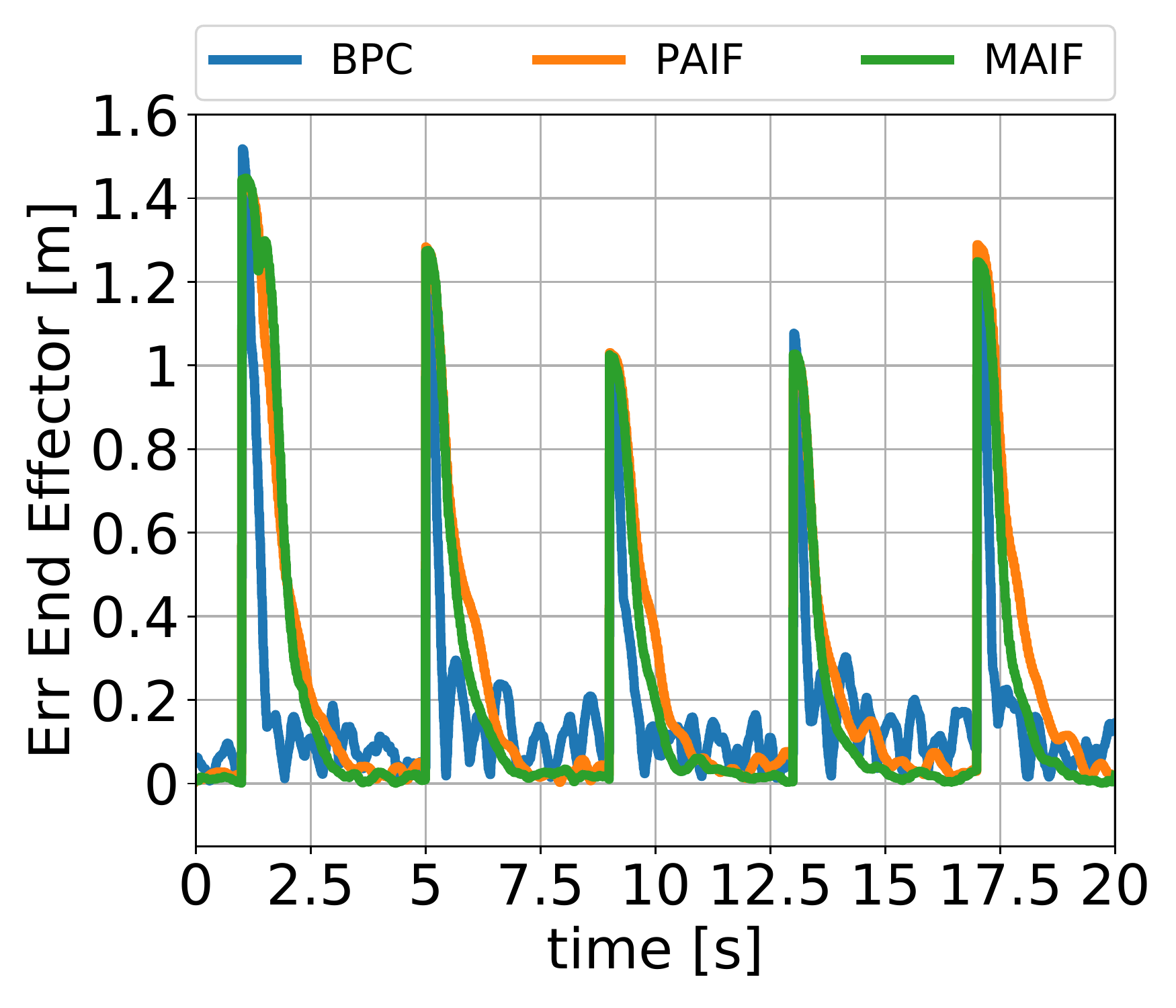}
    \caption{Sensory noise experiment. Sequential reaching of five goals under proprioceptive noise. All end-effector absolute errors present peaks when the new goal is set.}
    \label{fig:noisy_exp}
\end{figure}

We reevaluated the controller behavior in the presence of proprioceptive noise but in a sequential reaching task, focusing on the adaptation capabilities of the three controllers. Figure \ref{fig:noisy_exp} shows that MAIF was the most adaptive, presenting the smoothest behavior. PAIF also adapted to the noise but presented higher oscillations than MAIF. By contrast, BPC had the worst behavior, showing strong oscillations around the desired goals. 

\section{Conclusion}
\label{sec:conclusions}

We presented an improved adaptive torque controller for robot manipulators based on active inference. Our approach makes use of the alleged adaptability and robustness of AIF, taking advantage of previous works and overwhelming some related limitations. We solved state estimation by combining representation learning (with a multimodal VAE) with free energy optimization, improving the representational power and adaptability. As a result, we derived a schema for online proprio-visual torque control, which does not require any dynamic or kinematic model of the robot, is less sensitive to unmodeled dynamics, and can handle high-dimensional inputs. Although in this work we used images it can be generalized to any sensor modality. 

Results showed that our proposed algorithm outperforms in terms of adaptation the built-in Panda controller and a state-of-the-art torque AIF baseline. Moreover, it was more accurate in the presence of sensory noise. Our AIF controller was highly adaptive and robust to different contexts, such as sensory noise, visual occlusions, changes in the world dynamics (i.e., gravity) and changes in the robot properties (i.e. motor stiffness). Besides, our approach natively allowed mental simulation of the planned trajectory.

Future work will finalize the physical robot experiments that were not possible due to university restrictions and will investigate strongly coupled cross-modal learning and hierarchical schemes for sequential decision making.

\section*{Acknowledgment}
\addcontentsline{toc}{section}{Acknowledgment}
We would like to thank Professor Martijn Wisse for his fruitful comments on the elaboration of this work.


%
%
%

\bibliographystyle{IEEEtran}
\bibliography{references}

\begin{thebibliography}{10}
\providecommand{\url}[1]{#1}
\csname url@samestyle\endcsname
\providecommand{\newblock}{\relax}
\providecommand{\bibinfo}[2]{#2}
\providecommand{\BIBentrySTDinterwordspacing}{\spaceskip=0pt\relax}
\providecommand{\BIBentryALTinterwordstretchfactor}{4}
\providecommand{\BIBentryALTinterwordspacing}{\spaceskip=\fontdimen2\font plus
\BIBentryALTinterwordstretchfactor\fontdimen3\font minus
  \fontdimen4\font\relax}
\providecommand{\BIBforeignlanguage}[2]{{%
\expandafter\ifx\csname l@#1\endcsname\relax
\typeout{** WARNING: IEEEtran.bst: No hyphenation pattern has been}%
\typeout{** loaded for the language `#1'. Using the pattern for}%
\typeout{** the default language instead.}%
\else
\language=\csname l@#1\endcsname
\fi
#2}}
\providecommand{\BIBdecl}{\relax}
\BIBdecl

\bibitem{friston2010free}
K.~Friston, ``The free-energy principle: a unified brain theory?'' \emph{Nature
  reviews neuroscience}, vol.~11, no.~2, pp. 127--138, 2010.

\bibitem{friston2008dem}
F.~K.J, T.-B. N., and Daunizeau, ``Dem: a variational treatment of dynamic
  systems,'' \emph{NeuroImage, 41, pp. 849-885}, 2008.

\bibitem{sarkka2013bayesian}
S.~S{\"a}rkk{\"a}, \emph{Bayesian filtering and smoothing}.\hskip 1em plus
  0.5em minus 0.4em\relax Cambridge University Press, 2013, no.~3.

\bibitem{millidge2020relationship}
B.~Millidge, A.~Tschantz, A.~K. Seth, and C.~L. Buckley, ``On the relationship
  between active inference and control as inference,'' in \emph{International
  Workshop on Active Inference}.\hskip 1em plus 0.5em minus 0.4em\relax
  Springer, 2020, pp. 3--11.

\bibitem{lanillos2018adaptive}
P.~Lanillos and G.~Cheng, ``Adaptive robot body learning and estimation through
  predictive coding,'' in \emph{2018 IEEE/RSJ International Conference on
  Intelligent Robots and Systems (IROS)}.\hskip 1em plus 0.5em minus
  0.4em\relax IEEE, 2018, pp. 4083--4090.

\bibitem{oliver2021empirical}
G.~Oliver, P.~Lanillos, and G.~Cheng, ``An empirical study of active inference
  on a humanoid robot,'' \emph{IEEE Transactions on Cognitive and Developmental
  Systems}, 2021.

\bibitem{sancaktar2020end}
C.~Sancaktar, M.~A. van Gerven, and P.~Lanillos, ``End-to-end pixel-based deep
  active inference for body perception and action,'' in \emph{2020 Joint IEEE
  10th International Conference on Development and Learning and Epigenetic
  Robotics (ICDL-EpiRob)}.\hskip 1em plus 0.5em minus 0.4em\relax IEEE, 2020,
  pp. 1--8.

\bibitem{pezzato2020novel}
C.~Pezzato, R.~Ferrari, and C.~H. Corbato, ``A novel adaptive controller for
  robot manipulators based on active inference,'' \emph{IEEE Robotics and
  Automation Letters}, vol.~5, no.~2, pp. 2973--2980, 2020.

\bibitem{pio2016active}
L.~Pio-Lopez, A.~Nizard, K.~Friston, and G.~Pezzulo, ``Active inference and
  robot control: a case study,'' \emph{Journal of The Royal Society Interface},
  vol.~13, no. 122, p. 20160616, 2016.

\bibitem{minju2019goal}
M.~Jung, T.~Matsumoto, and J.~Tani, ``Goal-directed behavior under variational
  predictive coding: Dynamic organization of visual attention and working
  memory,'' \emph{IROS}, 2019.

\bibitem{baioumy2020active}
M.~Baioumy, P.~Duckworth, B.~Lacerda, and N.~Hawes, ``Active inference for
  integrated state-estimation, control, and learning,'' \emph{arXiv preprint
  arXiv:2005.05894}, 2020.

\bibitem{meera2020free}
A.~A. Meera and M.~Wisse, ``Free energy principle based state and input
  observer design for linear systems with colored noise,'' in \emph{2020
  American Control Conference (ACC)}.\hskip 1em plus 0.5em minus 0.4em\relax
  IEEE, 2020, pp. 5052--5058.

\bibitem{lanillos2020robot}
P.~Lanillos, J.~Pages, and G.~Cheng, ``Robot self/other distinction: active
  inference meets neural networks learning in a mirror,'' in \emph{Proceedings
  of the 24th European Conference on Artificial Intelligence (ECAI)}, 2020, pp.
  2410 -- 2416.

\bibitem{friston2010action}
K.~J. Friston, J.~Daunizeau, J.~Kilner, and S.~J. Kiebel, ``Action and
  behavior: a free-energy formulation,'' \emph{Biological cybernetics}, vol.
  102, no.~3, pp. 227--260, 2010.

\bibitem{Yuge2019Variational}
Y.~Shi, N.~Siddharth, B.~Paige, and P.~H. Torr, ``Variational
  mixture-of-experts autoencoders for multi-modal deep generative models,''
  \emph{NeurIPS}, 2019.

\bibitem{rood2020deep}
T.~Rood, M.~van Gerven, and P.~Lanillos, ``A deep active inference model of the
  rubber-hand illusion,'' in \emph{Active Inference}, T.~Verbelen, P.~Lanillos,
  C.~L. Buckley, and C.~De~Boom, Eds.\hskip 1em plus 0.5em minus 0.4em\relax
  Cham: Springer International Publishing, 2020, pp. 84--91.

\bibitem{lesort2018state}
T.~Lesort, N.~D{\'\i}az-Rodr{\'\i}guez, J.-F. Goudou, and D.~Filliat, ``State
  representation learning for control: An overview,'' \emph{Neural Networks},
  vol. 108, pp. 379--392, 2018.

\bibitem{friston2007variational}
K.~Friston, J.~Mattout, N.~Trujillo-Barreto, J.~Ashburner, and W.~Penny,
  ``\BIBforeignlanguage{English}{Variational free energy and the laplace
  approximation},'' \emph{\BIBforeignlanguage{English}{NeuroImage}}, vol.~34,
  no.~1, pp. 220--234, Jan. 2007.

\bibitem{buckley2017free}
C.~L. Buckley, C.~S. Kim, S.~McGregor, and A.~K. Seth, ``The free energy
  principle for action and perception: A mathematical review,'' \emph{Journal
  of Mathematical Psychology}, vol.~81, pp. 55--79, 2017.

\bibitem{otto2020deep}
O.~van~der Himst and P.~Lanillos, ``Deep active inference for partially
  observable mdps,'' in \emph{International Workshop on Active
  Inference}.\hskip 1em plus 0.5em minus 0.4em\relax Springer, 2020, pp.
  61--71.

\bibitem{lanillos2018active}
P.~Lanillos and G.~Cheng, ``Active inference with function learning for robot
  body perception,'' in \emph{Proc. Int. Workshop Continual Unsupervised
  Sensorimotor Learn.}, 2018, pp. 1--5.

\bibitem{franka}
\BIBentryALTinterwordspacing
``Franka emika panda robot arm.'' [Online]. Available:
  \url{https://www.franka.de/}
\BIBentrySTDinterwordspacing

\bibitem{koubaa2019ROS}
A.~Koubaa, \emph{Robot Operating System (ROS): The Complete Reference (Volume
  2)}, 1st~ed.\hskip 1em plus 0.5em minus 0.4em\relax Springer Publishing
  Company, Incorporated, 2017.

\bibitem{stevens2020pytorch}
A.~Paszke, S.~Gross, F.~Massa, A.~Lerer, J.~Bradbury, G.~Chanan, T.~Killeen,
  Z.~Lin, N.~Gimelshein, L.~Antiga, A.~Desmaison, A.~Kopf, E.~Yang, Z.~DeVito,
  M.~Raison, A.~Tejani, S.~Chilamkurthy, B.~Steiner, L.~Fang, J.~Bai, and
  S.~Chintala, ``Pytorch: An imperative style, high-performance deep learning
  library,'' in \emph{Advances in Neural Information Processing Systems 32},
  H.~Wallach, H.~Larochelle, A.~Beygelzimer, F.~d'~Alché-Buc, E.~Fox, and
  R.~Garnett, Eds.\hskip 1em plus 0.5em minus 0.4em\relax Curran Associates,
  Inc., 2019, pp. 8024--8035.

\bibitem{kingma2017adam}
D.~P. Kingma and J.~Ba, ``Adam: A method for stochastic optimization,'' 2017.

\end{thebibliography}

\end{document}